\documentclass[10pt,journal,compsoc]{IEEEtran}
\usepackage{graphicx}
\usepackage{xcolor}
\usepackage{tikz}
\usepackage{comment}
\usepackage{amsmath,amssymb} % define this before the line numbering.
\usepackage{color}
\usepackage{graphicx}
\usepackage{amsmath}
\usepackage{amssymb}
\usepackage{booktabs}

\usepackage{algorithm}
\usepackage{algorithmic}

\usepackage{colortbl}
\usepackage{amssymb}
\usepackage{amsmath}
\usepackage{multirow}
\usepackage{newfloat}
\usepackage{listings}
\usepackage{color}
\usepackage{bbding}
\usepackage{amsmath}
\usepackage{amsthm}

\newtheorem{proposition}{Proposition}
\ifCLASSOPTIONcompsoc
  \usepackage[nocompress]{cite}
\else
  \usepackage{cite}
\fi
\ifCLASSINFOpdf
\else
\fi
\begin{document}
\title{Improving Fast Adversarial Training with Prior-Guided Knowledge}

\author{Xiaojun Jia,~\IEEEmembership{Student Member,~IEEE,}
        Yong Zhang,
        Xingxing Wei,
        Baoyuan Wu,
        Ke Ma, \\
        Jue Wang~\IEEEmembership{Fellow,~IEEE,} 
        and~Xiaochun Cao~\IEEEmembership{Senior Member,~IEEE}% <-this % stops a space
\IEEEcompsocitemizethanks{
\IEEEcompsocthanksitem Xiaojun Jia is with
State Key Laboratory of Information Security,
Institute of Information Engineering, Chinese Academy of Sciences, Beijing 100093, China, and also with School of Cyber Security, University of Chinese Academy of Sciences, Beijing 100049, China.
(e-mail: jiaxiaojun@iie.ac.cn)
% \IEEEcompsocthanksitem Yang Bai is with Tsinghua University, Beijing 100084, China. y-bai17@mails.tsinghua.edu.cn
\IEEEcompsocthanksitem Yong Zhang and Jue Wang are with AI Lab, Tencent Inc., Shenzhen 518000, China.(e-mail:
  $\{$zhangyong201303, arphid$\}$@gmail.com)
\IEEEcompsocthanksitem Xingxing Wei is with Institute of Artificial Intelligence, Beihang University, Beijing, 100191, P.R. China.(e-mail:xxwei@buaa.edu.cn) 
\IEEEcompsocthanksitem Baoyuan Wu is with School of Data Science, the Chinese University of Hong Kong, Shenzhen (CUHK-Shenzhen) and Secure Computing Lab of Big Data, Shenzhen Research Institute of Big Data (SBRID), Shenzhen 518172, China. 
(e-mail: wubaoyuan@cuhk.edu.cn)
\IEEEcompsocthanksitem Ke Ma is with the School of Electronic, Electrical and Communication Engineering, University of Chinese Academy of Sciences, Beijing 100049, China. (E-mail: make@ucas.ac.cn)
\IEEEcompsocthanksitem  Xiaochun Cao (Corresponding) is with School of Cyber Science and Technology, Shenzhen Campus, Sun Yat-sen University, Shenzhen 518107, China (e-mail: caoxiaochun@mail.sysu.edu.cn)}% <-this % stops a space
% \thanks{Manuscript received April 19, 2005; revised August 26, 2015.}
}


\markboth{IEEE Transactions on Pattern Analysis and Machine Intelligence}%
{Shell \MakeLowercase{\textit{et al.}}: Bare Advanced Demo of IEEEtran.cls for IEEE Computer Society Journals}



\IEEEtitleabstractindextext{%

\begin{abstract}
Fast adversarial training (FAT) is an efficient method to improve robustness. However, the original FAT suffers from catastrophic overfitting, which dramatically and suddenly reduces robustness after a few training epochs. Although various FAT variants have been proposed to prevent overfitting, they require high training costs. In this paper, we investigate the relationship between adversarial example quality and catastrophic overfitting by comparing the training processes of standard adversarial training and FAT. We find that catastrophic overfitting occurs when the attack success rate of adversarial examples becomes worse. Based on this observation, we propose a positive prior-guided adversarial initialization to prevent overfitting by improving adversarial example quality without extra training costs. This initialization is generated by using high-quality adversarial perturbations from the historical training process. We provide theoretical analysis for the proposed initialization and propose a prior-guided regularization method that boosts the smoothness of the loss function. Additionally, we design a prior-guided ensemble FAT method that averages the different model weights of historical models using different decay rates. Our proposed method, called FGSM-PGK, assembles the prior-guided knowledge, \emph{i.e.,} the prior-guided initialization and model weights, acquired during the historical training process. Evaluations of four datasets demonstrate the superiority of the proposed method.

\end{abstract}


% \begin{abstract}
% The abstract goes here.
% \end{abstract}

% % Note that keywords are not normally used for peerreview papers.
% \begin{IEEEkeywords}
% Computer Society, IEEE, IEEEtran, journal, \LaTeX, paper, template.
% \end{IEEEkeywords}

}

\maketitle

% is not selected <OR> if conference mode is selected - because compsoc
% conference papers position the abstract like regular (non-compsoc)
\IEEEdisplaynontitleabstractindextext

% \fi
\IEEEpeerreviewmaketitle

\section{Introduction}
\label{sec:introduction}
\IEEEPARstart{D}{eep} neural networks (DNNs)~\cite{simonyan2014very,he2016deep,zagoruyko2016wide} have achieved cutting-edge performance in a lot of problems and tasks. However, they have been shown to be vulnerable to adversarial perturbations that are often imperceptible to human observers~\cite{szegedy2013intriguing,goodfellow2014explaining}. This vulnerability poses a significant security risk for sensitive applications of DNNs. To address this issue, a great deal of defense methods~\cite{song2017pixeldefend,jia2019comdefend,yin2020defense,dai2022deep} have been developed to improve the adversarial robustness of DNNs against adversarial attacks. Standard adversarial training~\cite{madry2017towards,wang2019improving,rebuffi2021fixing,jia2022adversarial} has been demonstrated as one of the most effective methods to enhance the robustness of DNNs against adversarial examples. However, most of them adopt a multi-step adversarial attack, \emph{i.e.,} Projected Gradient Descent (PGD)~\cite{madry2017towards}, to generate adversarial examples for training. This approach incurs a significant training cost, which limits the practical application of standard adversarial training. 
\begin{figure}[t]

        \centering
        \includegraphics[width=0.9\linewidth]{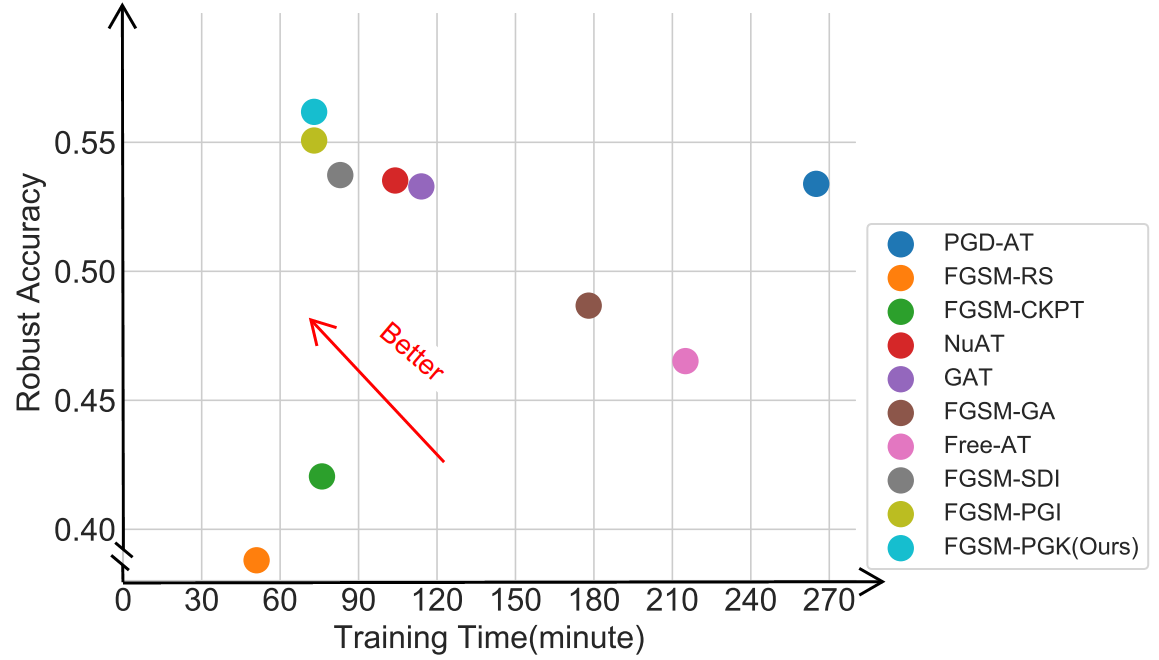}
        \label{fig:time_acc}

  \caption{ PGD-10 accuracy and training time of various fast adversarial training methods with ResNet18 as the backbone on the CIFAR-10 dateset. The $x$-axis represents training time (lower values indicate higher efficiency) and the $y$-axis represents PGD-10 accuracy (higher values indicate greater robustness).  }
\label{fig:time_acc}
\end{figure}
\begin{figure*}[t]
\centering
 \includegraphics[width=0.95\linewidth]{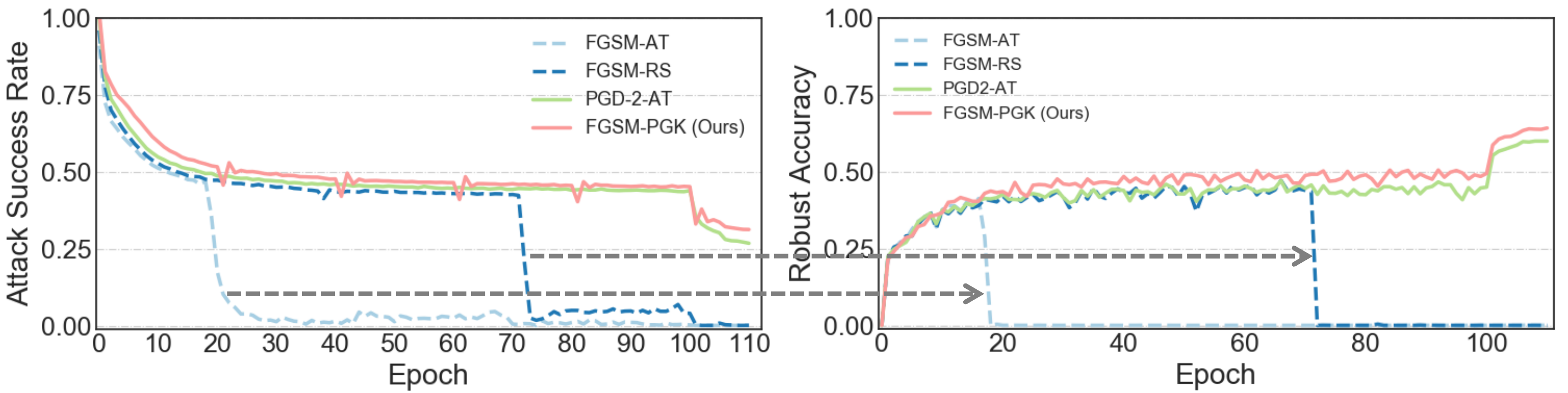}

  \caption{
  The difference between the training processes of fast adversarial training and standard adversarial training methods.
  The left figure shows the attack success rate of generated adversarial examples, and the right figure shows the PGD-10 robust accuracy of the target model.
It is observed that FGSM-AT and FGSM-RS experience catastrophic overfitting after training for a few epochs, but PGD-2-AT and our proposed method effectively prevent overfitting.
  %The attack success rate of AT methods on the CIFAR10 database in the training phase. 
%   As the training progresses, the model gradually becomes robust, and the attack success rate of generated adversarial examples gradually decreases. 
%   The PGD-10 accuracy of AT methods on the CIFAR10 database in the training phase. As the training progresses, the model gradually becomes robust, and the PGD-10 accuracy gradually increases. But the FGSM-AT and FGSM-RS encounter catastrophic  overfitting.
  }
\label{fig:overfitting}
\end{figure*}
\par To reduce training costs, fast adversarial training~\cite{tramer2017ensemble}, which can be formulated as a mini-max optimization problem,  has been proposed to improve adversarial robustness.  During the inner maximization step, a single-step adversarial attack, such as the Fast Gradient Sign Method (FGSM), is used to generate adversarial examples for training. However, after just a few training epochs, the model suddenly loses its robustness. This phenomenon is described as  
catastrophic overfitting~\cite{wong2020fast}. To address this issue, several fast adversarial training variants~\cite{andriushchenko2020understanding,sriramanan2020guided,kim2021understanding} have been proposed from different perspectives to enhance the quality of adversarial examples and prevent overfitting. These approaches can be broadly categorized as sample initialization~\cite{jia2022boosting} and regularization~\cite{sriramanan2021towards}. They not only relieve catastrophic overfitting but also achieve cutting-edge model robustness performance. However, compared with the original FGSM-AT~\cite{tramer2017ensemble}, they require extra high training costs to improve the quality of adversarial examples. 
\par The quality of adversarial examples is critical in fast adversarial training. In this study, we explore the difference in the quality of adversarial examples between fast adversarial training and standard adversarial training to understand the reason for catastrophic overfitting. Surprisingly, we observe that the attack success rate of adversarial examples is significantly different between fast adversarial training and standard adversarial training after a few training epochs. Specifically, as shown in Fig.~\ref{fig:overfitting}, after a certain period of training, the attack success rate of adversarial examples used in fast adversarial training methods (FGSM-AT and FGSM-RS) suddenly and sharply drops, and at the same time, the adversarial robustness of the model also drops sharply. This finding indicates that catastrophic overfitting is related to the quality of the adversarial examples, \emph{i.e.}, fast adversarial training meets catastrophic overfitting when the quality of the adversarial examples deteriorates. However, PGD-2-AT (two-step PGD-AT), which can be considered as FGSM-AT with adversarial sample initialization, does not meet catastrophic overfitting. This result suggests that better sample initialization could help fast adversarial training prevent catastrophic overfitting.

\par Based on above observations, we raise the question of whether it is possible to obtain an adversarial sample initialization without incurring extra training costs to improve the quality of adversarial examples and prevent catastrophic overfitting in fast adversarial training.
We investigate some sample initialization strategies for fast adversarial training and find that better initialization could improve the adversarial example quality to prevent catastrophic overfitting. Therefore, we propose to adopt the prior-guided initialization which is generated by utilizing the high-quality adversarial perturbations from the historical training process. % \emph{dubbed} FGSM-PGI. Specifically, we use buffered adversarial perturbations from the previous batch and epoch to initialize samples for fast adversarial training, \emph{dubbed} as FGSM-BP and FGSM-EP, respectively. These two sample initialization strategies not only prevent catastrophic overfitting but also improve model robustness. To make better use of prior information, 
 Specifically, we propose to use the buffer gradient from all previous periods as an additional prior through the momentum mechanism, and furthermore, %, \emph{dubbed} as FGSM-MEP. \textcolor{red}{Additionally, 
 accumulate the buffer gradient with different weights based on the quality of historical adversarial examples. 

We also integrate the above prior-guided sample initialization into the minimization process in fast adversarial training and propose a simple yet effective regularization method to further enhance model robustness. Specifically, the proposed regularization method is used to prevent the output of the learning model on the current adversarial examples from deviating too much from the output on the samples initialized by the prior-guided initialization. In the optimization step of minimizing, we minimize the squared $\mathcal{L}_{2}$ distance between the model predictions on the two types of adversarial examples generated by prior-guided initialization and adversarial perturbations. The proposed regularization term boosts the smoothness of loss function around samples by forcing the learned model to be robust to the two types of adversarial examples. 

% We adopt a squared $L_{2}$ distance to implement the proposed regularization method. \textcolor{blue}{I suggest to add some texts to detail this module.}

\par 
In addition,  model weight averaging (WA)~\cite{izmailov2018averaging} has been proposed previously to improve model generalization by accumulating historical model weights during the training process, with no additional computational overhead. 
% which is implemented by accumulating the historical model weights during the training process, has been proposed to improve the model generalization with no computational overhead. 
Previous works ~\cite{gowal2020uncovering,chen2020robust} have shown that WA can significantly improve model robustness. Several recent research \cite{rebuffi2021fixing,wang2022self} has also suggested that incorporating WA can increase the robustness of the standard adversarial training. In this paper, we investigate the influence of WA on fast adversarial training and find that directly using WA results in limited adversarial robustness improvement.  This is because non-robust model weights are also present in the fast adversarial training process, which can negatively affect the final model's robustness. To overcome this limitation, we propose a prior-guided ensemble fast adversarial training method that averages the model weights of historical models with different decay rates during the training process.

\par  
 By assembling the proposed prior-guided knowledge, \emph{i.e.,} the prior-guided initialization and model weights, we conclude our fast adversarial training method, \emph{dubbed} FGSM-PGK. The proposed method is capable of not only effectively preventing catastrophic overfitting but also achieving cutting-edge adversarial robustness, surpassing the performance of state-of-the-art fast adversarial training methods. Our main contributions are in five aspects:
\begin{itemize}
    \item  We explore several initialization strategies for fast adversarial training and discover that superior initialization can prevent catastrophic overfitting. Furthermore, we provide a theoretical analysis to support our findings. 
    \item We propose a prior-guided initialization with different weights to conduct fast adversarial training to improve model robustness. 
    \item We propose a simple yet effective regularization method for the prior-guided initialization to improve model robustness.
    \item We propose a prior-guided ensemble fast adversarial training method that averages the model weights of historical models with different decay rates to improve adversarial robustness.
    \item Our extensive experiments across various network architectures and datasets demonstrate that the proposed method outperforms state-of-the-art fast adversarial training methods with lower additional training costs. 
\end{itemize}

\par This paper is a journal extension of our conference paper~\cite{jia2022prior} (called FGSM-PGI). We have made significant improvements and extensions in this version compared to the preliminary conference version. The main differences are in four aspects: 1) In addition to the prior-guided knowledge of initialization proposed in the previous version, we also consider another form of such knowledge: the model weights. We regard model weights as a form of prior-guided knowledge and investigate the impact of weight averaging on fast adversarial training. Our findings suggest that using weight averaging directly only leads to limited improvements in adversarial robustness due to the presence of non-robust model weights in fast adversarial training. 2) To overcome this limitation, we propose a prior-guided ensemble fast adversarial training method that averages the model weights of historical models with different decay rates in Section \ref{Proposed_weights}. Moreover, we also propose to accumulate the buffer gradient with different weights based on the quality of historical adversarial examples in Section \ref{Proposed_Initialization} to prevent catastrophic overfitting and achieve better adversarial robustness. 3) More experiments and analyses are given and discussed involving comparisons with SOTA methods, ablation study, and performance analysis. Specifically, we add the ablation study versus different prior-guided elements in Section \ref{Ablation_Study}, and analysis the effectiveness of performance in Section \ref{Performance_Analysis}. One SOTA fast adversarial training method recently published in TIP2022 is also added as the new comparisons in Section \ref{Comparison_Experiments}. 4) We completely rewrite the abstract, introduction, method, experiment, and conclusion sections to provide a more comprehensive overview of our motivation and approach. Additionally, we revamp all the figures and tables.

\par The remainder of this paper is structured as follows. We briefly
review related works, including adversarial attacks and adversarial training in Section \ref{sec:related_work}. In Section \ref{sec:proposed_method}, we
rethink catastrophic overfitting and delve into the underlying reasons for the occurrence of catastrophic overfitting. Thus, we introduce the details of
the proposed method. In Section~\ref{sec:experiments}, we present the results of extensive experiments and provide relevant analysis. Finally, Section \ref{sec:conclusion} concludes the paper, and we hope that our work can inspire a deeper understanding of fast adversarial training and contribute to improving the adversarial robustness of DNNs.
\section{Related Work}
\label{sec:related_work}
\subsection{Adversarial attack}
Deep neural networks (DNNs) have been found to be vulnerable to adversarial examples, which are generated by adding imperceptible adversarial perturbations~\cite{szegedy2013intriguing}. A series of works have been proposed to improve the performance of the adversarial attack~\cite{jia2020adv, bai2020improving, dong2021query,wei2022simultaneously,dong2022viewfool,yang2022boosting}. In detail, Goodfellow \emph{et al.} ~\cite{goodfellow2014explaining} propose the Fast Gradient Sign Method (FGSM) adversarial attack method, which generates adversarial perturbation in the
direction of the gradient of the loss function. Papernot \emph{et al.}~\cite{papernot2016limitations}
propose a stronger adversarial attack method, which only modifies a few input pixels based on a Jacobian-based saliency map. Madry \emph{et al.}~\cite{madry2017towards} propose a Projected Gradient Decent (PGD) adversarial attack method, which is a multi-step optimal first-order attack method. Moosavi \emph{et al.}~\cite{moosavi2016deepfool} propose a non-targeted adversarial attack method, which generates adversarial examples by finding the closest decision boundary of the DNNs, called Deepfool. 
Carlini and Wagner~\cite{carlini2017towards} propose a powerful adversarial attack method, which can overcome several pre-processing defense methods. 
Besides, Croce \emph{et al.}~\cite{croce2020reliable}
propose two parameter-free attack methods, which are the improved versions of PGD-attack, called APGD-CE and APGD-DLR. They do not require a hand-crafted step size of the loss function. Thus, the proposed two methods are combined with two other adversarial attack methods (Square~\cite{andriushchenko2020square} and FAB~\cite{croce2020minimally}) to form a more powerful attack method that is wide to evaluate the adversarial robustness, called AutoAttack (AA).

\subsection{Adversarial training}
Adversarial training has been demonstrated as one of the most effective approaches to improve adversarial robustness, which can be formulated as a mini-max optimization problem as follows:
% \begin{equation}
% \begin{gathered}
% \underset{\boldsymbol{w}}{\arg \min } \mathbb{E}_{(\boldsymbol{x}, y) \sim \mathcal{D}}\left[\max _\delta \mathcal{L}\left(f_{\boldsymbol{w}}(\boldsymbol{x}+\boldsymbol{\delta}), y\right)\right] \\
% \text { s.t. }\|\boldsymbol{\delta}\|_p \leq \epsilon, 
% \end{gathered}
% \end{equation}
\begin{equation}
    \begin{aligned}
        & &\underset{\boldsymbol{w}}{\textbf{\textit{arg min}}}&\ \ \mathbb{E}_{(\boldsymbol{x}, y) \sim \mathcal{D}}\left[\underset{\delta}{\textbf{\textit{max}}}\   \mathcal{L}\left(f_{\boldsymbol{w}}(\boldsymbol{x}+\boldsymbol{\delta}), y\right)\right] \\[3pt]
        & &\textbf{\textit{s.t.}}&\ \ \|\boldsymbol{\delta}\|_p \leq \epsilon, 
    \end{aligned}
\end{equation}
where $f_{\boldsymbol{w}}(\cdot)$ represents DNN with weights $\boldsymbol{w}$, $\mathcal{D}$ is a data distribution with the benign sample $\boldsymbol{x}$ and the corresponding ground truth label $y$, $\boldsymbol{\delta}$ is the generated adversarial perturbation, $\epsilon$ is the maximum perturbation strength and $\mathcal{L}\left(f_{\boldsymbol{w}}(\boldsymbol{x}), y\right)$ represents the cross entropy loss function.
Standard adversarial training methods always adopt a multi-step adversarial attack, \emph{i.e.,} Projected Gradient Descent (PGD)~\cite{madry2017towards}, to generate adversarial examples to improve adversarial robustness. The adversarial perturbation can be defined as:
% \begin{equation}
% \boldsymbol{\delta}_{t}=\Pi_{[-\epsilon, \epsilon]}\left[\boldsymbol{\delta}_{t-1}+\alpha \cdot \operatorname{sign}\left(\nabla_{\boldsymbol{x}} \mathcal{L}\left(f_{\boldsymbol{w}}\left(\boldsymbol{x}+\boldsymbol{\delta}_{t-1} \right), y\right)\right)\right],
% \end{equation}
\begin{equation}
    \begin{aligned}
        \boldsymbol{\delta}_{t} = \prod_{[-\epsilon, \epsilon]}\Big[\boldsymbol{\delta}_{t-1}+\alpha \cdot \operatorname{sign}\left(\nabla_{\boldsymbol{x}} \mathcal{L}\left(f_{\boldsymbol{w}}\left(\boldsymbol{x}+\boldsymbol{\delta}_{t-1} \right), y\right)\right)\Big],
    \end{aligned}
\end{equation}
where $\boldsymbol{\delta}_{t}$ is the adversarial perturbation for $t$ iterations and $\alpha$ is the step size. Madry \emph{et al.}~\cite{madry2017towards} first adopt PGD to conduct adversarial training and propose a prime PGD-based AT framework. After that, a series of standard adversarial training variants ~\cite{kannan2018adversarial,wang2019improving,pang2020boosting,roth2020adversarial,bai2021clustering,bai2021improving,jia2022adversarial,huq2023mixpgd,wu2022towards,li2023recognizing} have been proposed to significantly improve adversarial robustness from different perspectives. A sample yet effective PGD-based AT variant~\cite{rice2020overfitting}, \emph{i.e.,} PGD-AT with an early stopping strategy,  stands out amongst them. But they calculate the gradients multiple times which requires plenty of training costs. It limits the practical application of standard adversarial training. To improve the training efficiency, Goodfellow \emph{et al.}~\cite{goodfellow2014explaining} propose to adopt a single-step adversarial attack, \emph{i.e.,} FGSM, to generate adversarial examples for training, \emph{called} FGSM-AT. The adversarial perturbation for FGSM-AT can be defined as:
\begin{equation}
\mathbf{\boldsymbol{\delta}}=\alpha \cdot \textrm{sign}\left(\nabla_{\boldsymbol{x}} \mathcal{L}(f_{\boldsymbol{w}}(\boldsymbol{x}), y)\right),
\label{Eq:FGSM_AT}
\end{equation}
where  $\alpha$ is set to $\epsilon$. Wong \emph{et al.}~\cite{wong2020fast} have indicated that although FGSM-AT reduces the training costs for adversarial training, it could suffer severe catastrophic overfitting, \emph{i.e.,} after a few training epochs, FGSM-AT could suddenly lose the adversarial robustness. To alleviate catastrophic overfitting, they propose to initialize the samples with a random initialization for FGSM-AT, called FGSM-RS. The adversarial perturbation for FGSM-RS can be defined as:
% \begin{equation}
% \mathbf{\boldsymbol{\delta}}=\Pi_{[-\epsilon, \epsilon]}\left[\mathbf{\boldsymbol{\boldsymbol{\eta}}}+\alpha \cdot \textrm{sign}\left(\nabla_{\boldsymbol{x}} \mathcal{L}(f_{\boldsymbol{w}}(\boldsymbol{x}+\mathbf{\boldsymbol{\boldsymbol{\eta}}}), y)\right)\right],
% \label{eq:FGSM_RS}
% \end{equation}
\begin{equation}
    \begin{aligned}
        \mathbf{\boldsymbol{\delta}}=\prod_{[-\epsilon, \epsilon]}\Big[\mathbf{\boldsymbol{\boldsymbol{\eta}}}+\alpha \cdot \textrm{sign}\left(\nabla_{\boldsymbol{x}} \mathcal{L}(f_{\boldsymbol{w}}(\boldsymbol{x}+\mathbf{\boldsymbol{\boldsymbol{\eta}}}), y)\right)\Big],
    \end{aligned}
\end{equation}
where $\boldsymbol{\eta}$ is the random initialization, which belongs to a uniform distribution $\boldsymbol{U}(-\epsilon, \epsilon)$, and $\alpha$ is set to $1.25\epsilon$. But some research~\cite{andriushchenko2020understanding,kim2021understanding} has claimed that after training for a longer period of time, FGSM-RS could also encounter catastrophic overfitting. 
After that, several fast adversarial training variants ~\cite{andriushchenko2020understanding,kim2021understanding,jia2022boosting,li2022subspace} have been proposed to 
prevent catastrophic overfitting and further improve adversarial robustness by enhancing the adversarial example quality from different perspectives, such as sample initialization and regularization.
Andriushchenko \emph{et al.} \cite{andriushchenko2020understanding} propose a  regularization method based on gradient alignment to improve the quality of adversarial examples for training, called FGSM-GA. It can be defined as:
\begin{equation}
Reg=1-\cos \left(\nabla_{\boldsymbol{x}} \mathcal{L}(f_{\boldsymbol{w}}(\boldsymbol{x}), {y}), \nabla_{\boldsymbol{x}} \mathcal{L}(f_{\boldsymbol{w}}(\boldsymbol{x}+\boldsymbol{\eta}), {y})\right).
\end{equation}
Sriramanan \emph{et al.} \cite{sriramanan2020guided} propose a function smoothing guided regularization to improve adversarial robustness, called GAT. It can be defined as:
\begin{equation}
 Reg=\left\|f_{\boldsymbol{w}}(\boldsymbol{x}+\boldsymbol{\delta}) -f_{\boldsymbol{w}}(\mathbf{x})\right\|_2^2.
\end{equation}
Sriramanan \emph{et al.} \cite{sriramanan2021towards} propose a regularization method based on Nuclear-Norm to boost function smoothing, called NuAT. It can be defined as:
\begin{equation}
 Reg = \left\|f_{\boldsymbol{w}}(\boldsymbol{x}+\boldsymbol{\delta}) -f_{\boldsymbol{w}}(\boldsymbol{x} )\right\|_*,
\end{equation}
where  $\|\cdot\|_*$ is the Nuclear Norm function. Moreover, Jia \emph{et al.} \cite{jia2022boosting} propose to make use of a generative model to generate a learnable sample initialization to conduct fast adversarial training, called FGSM-SDI. They not only prevent catastrophic overfitting but also greatly improve adversarial robustness.  But they require much extra training costs to perform adversarial training. 
\section{The Proposed Method}
\label{sec:proposed_method}
In this section, our observations of rethinking catastrophic overfitting are presented in Sec.~\ref{Rethinking_CO}. The results of FGSM-AT combined with several sample initialization strategies are shown in Sec.~\ref{Initialization}. We propose a prior-guided sample initialization with different weights to boost fast adversarial training in Sec.~\ref{Proposed_Initialization}. Thus we propose a simple yet effective regularization method to guide model training in Sec.~\ref{Proposed_Regularization}. Moreover, we also design a prior-guided ensemble fast adversarial training with the positive model weights of the history models in Sec.~\ref{Proposed_weights}.

\subsection{Rethinking Catastrophic Overfitting}
\label{Rethinking_CO}
Wong \emph{et al.} \cite{wong2020fast} discover fast adversarial training could encounter catastrophic overfitting after a few training epochs, \emph{i.e.,} fast adversarial training suddenly losses robust accuracy on the adversarial examples during the late stages of training. Thus FGSM-AT with the random initialization (FGSM-RS) can prevent catastrophic overfitting with the limited training epochs. Kim \emph{i.e.,} \cite{kim2021understanding} indicate FGSM-RS trained for more training epochs could still encounter catastrophic overfitting and propose to find the optimal attack step size of FGSM-RS to improve the adversarial example quality (FGSM-CKPT) during the different stages of fast adversarial training. To prevent catastrophic overfitting, 
Andriushchenko \emph{et al.}  \cite{andriushchenko2020understanding} explicitly maximize the proposed gradient alignment regularization to improve the adversarial example quality. Sriramanan \emph{et al.} \cite{sriramanan2020guided}
make use of the maximum-margin loss term in the generation process of adversarial examples to generate stronger adversarial examples. Besides,  Sriramanan \emph{et al.} \cite{sriramanan2021towards} also add the regularization loss term based on the Nuclear Norm to the cross-entropy loss to improve the quality of adversarial examples. Moreover, Jia \emph{et al.} \cite{jia2022boosting} propose to adopt a generative network to boost the quality of adversarial examples for training. These methods potentially indicate that catastrophic overfitting is related to the adversarial example quality and improve the adversarial example quality to prevent catastrophic overfitting. But they require more training costs. In detail, FGSM-GA introduces a heavy computation overhead by applying alignment regularization on the gradient. FGSM-CKPT finds the optimal attack step size by performing forwarding propagation multiple times. GAT and NuAT implement their proposed regularization methods in the inner maximization of the loss function to generate strong adversarial examples. And FGSM-SDI requires extra training costs to train the generative network, which greatly reduces training efficiency. It is necessary to improve the adversarial example quality without extra training costs. 

\par We compare the intermediate properties of the fast and standard adversarial training during the different training stages to reinvestigate catastrophic overfitting. In detail, as for fast adversarial training, we adopt FGSM-RS and FGSM-AT which could severely suffer catastrophic overfitting to conduct experiments. 
As for standard adversarial training, we adopt PGD-2-AT which uses the PGD attack with two iterations to conduct experiments. The core difference between adversarial training and 
conventional training is that an inter maximization optimization problem exists in adversarial training, \emph{i.e.,} the generation of adversarial examples. We use the attack success rate of the adversarial examples to evaluate the quality of the adversarial examples and observe the adversarial example quality in the whole training. The curves of the attack success rate and robust accuracy across several benchmark datasets are shown in Fig.~\ref{fig:overfitting}. 

\par Our observations are summarized as follows. 
1) Firstly, we observe that after several training periods, the attack success rates of the  
FGSM-RS and FGSM-AT have deteriorated dramatically, which results in a sharp decline in robust accuracy. It indicates that if the adversarial examples used for training can not attack the learning model, the model could lose the adversarial robustness against adversarial examples. Note that we are the first one to investigate catastrophic overfitting from the perspective of the adversarial example quality, which is overlooked by the previous research. 2) Secondly, using random initialization can delay the occurrence of catastrophic overfitting. In detail, on CIFAR-10, 
catastrophic overfitting by using random initialization is postponed to a later training epoch (20-th epoch $\Rightarrow$ 70-th epoch).  Simply implementing the random initialization can only alleviate catastrophic overfitting, but can not fundamentally prevent it. 3) Thirdly, to our surprise, PGD-2-AT has never encountered catastrophic overfitting during the whole training, which attracts our attention. We can regard the PGD-2-AT as the FGSM-AT with the sample initialization generated by the FGSM attack. It can be observed that several high-quality adversarial examples, which can successfully attack the model,  still exist in the late stage of training. 
It indicates that using better sample initialization can achieve higher-quality adversarial examples. But it requires extra training costs, which is undesirable for fast adversarial training. 
\par   Rethink that the adversarial perturbation generated by the FGSM attack ~\cite{goodfellow2014explaining} can be defined as: 
% $\delta_{adv}=\arg \max _{\|\delta\|_{p}<\varepsilon}\left\langle\nabla_{\boldsymbol{x}} \mathcal{L}(f_{\boldsymbol{w}}(\boldsymbol{x}+\boldsymbol{\delta}),y), \delta\right\rangle$, 
\begin{equation}
    \begin{aligned}
        \boldsymbol{\delta}_{adv}=\underset{\|\delta\|_{p}<\varepsilon}{\textbf{\textit{arg max}}}\left\langle\nabla_{\boldsymbol{x}} \mathcal{L}(f_{\boldsymbol{w}}(\boldsymbol{x}+\boldsymbol{\delta}),y), \boldsymbol{\delta}\right\rangle,
    \end{aligned}
\end{equation}
where ${p}$ represents the constrained normal form $\mathcal{L}_{\infty}$. If the loss function has the local linear property, \emph{i.e.,} $\nabla_{\boldsymbol{x}} \mathcal{L}(f_{\boldsymbol{w}}(\boldsymbol{x}+\boldsymbol{\delta}),y)$ is constant under the $\mathcal{L}_{\infty}$, using a one-step FGSM attack could find the optimal solution of adversarial perturbation. Otherwise, the FGSM attack could not find the first-best adversary, which could result in catastrophic overfitting, but a multi-step PGD attack with extra training costs could find the first-best adversarial perturbation. It indicates that catastrophic overfitting is directly related to the quality of the adversarial examples. We can understand the effectiveness of fast adversarial training variants from this perspective. They adopt sample initialization and regularization methods to improve the solution of the inner maximization solution, \emph{i.e.,} the high-quality adversarial examples. Most of them require additional training costs. In this paper, we explore the sample initialization strategies and propose a prior-guided initialization with different weights to improve the quality of the adversarial examples without extra training costs.

\subsection{Sample Initialization Strategies}
\label{Initialization}
Motivated by the above observations, we delve into the problem of \textit{``How to obtain efficient adversarial initialization without increasing the computational cost?"}.
Inspired by the adversarial example transferability~\cite{li2020learning}, we excogitate to use the adversarial perturbations generated during the historical training process to initialize the FGSM-based adversarial examples for training. Such initialization perturbations can be considered as prior knowledge of the sample. They can be freely obtained without increasing the computational cost on the gradient calculation except for additional memory storage. In detail, we study three sample initialization strategies of taking advantage of prior-guided adversarial perturbations, \emph{i.e.,} adopting the sample initialization from the adversarial perturbations in the previous training batch, in the previous training epoch, and in the momentum of all training epochs. 

\subsubsection{Prior-guided Initialization From Previous Batch} The adversarial perturbations from the previous training batch are stored and used to initialize the FGSM-based adversarial examples in the current batch, \emph{dubbed} FGSM-BP. In each iteration, we obtain the training batch by randomly sampling from the training dataset. Hence, the adversarial perturbations in the previous batch have no correspondence with the samples in the current batch. In detail, as for the current benign image $\boldsymbol{x}$,
we perform FGSM on the perturbed example which is generated by adding the perturbation in the previous batch on $\boldsymbol{x}$. The adversarial perturbation can be calculated as:
\begin{equation}
\boldsymbol{\delta}_{B_{t}}=\prod_{[-\epsilon, \epsilon]}\Big[\boldsymbol{\delta}_{B_{t-1}}+\alpha\cdot\operatorname{sign}\left( \nabla_{\boldsymbol{x}} \mathcal{L}(f_{\boldsymbol{w}}(\boldsymbol{x}+\boldsymbol{\delta}_{B_{t-1}}), y)\right)\Big],
\label{eq:FSGM_BP}
\end{equation}
where $\boldsymbol{\delta}_{B_{t}}$ represents the perturbation in the batch at the $t$-th iteration. 

\subsubsection{Prior-guided Initialization From Previous Epoch}
All adversarial perturbations from the previous epoch are stored and used to initialize the FGSM-based adversarial examples in the current epoch. 
Different from FGSM-BP, the adversarial perturbations in the previous epoch are directly related to the samples in the current epoch. Specifically, as for the current benign image $\boldsymbol{x}$, we adopt the 
corresponding perturbation $\boldsymbol{\delta}_{E_{t-1}}$ from the previous epoch to initialize $\boldsymbol{x}$ and thus perform FGSM to generate adversarial examples for training. The adversarial perturbation can be calculated as:
\begin{equation}
\boldsymbol{\delta}_{E_{t}}=\prod_{[-\epsilon, \epsilon]}\Big[\boldsymbol{\delta}_{E_{t-1}}+\alpha \cdot \operatorname{sign}\left( \nabla_{\boldsymbol{x}} \mathcal{L}(f_{\boldsymbol{w}}(\boldsymbol{x}+\boldsymbol{\delta}_{E_{t-1}}), y)\right)\Big].
\label{eq:FSGM_EP}
\end{equation}

\subsubsection{Prior-guided initialization From the Momentum of All Previous Epochs}
 To make full use of the prior-guided adversarial perturbations, we propose to accumulate the gradient momentum information of one sample across all previous epochs to generate the sample initialization in the current epoch for FGSM-based adversarial example generation, called FGSM-MEP. In detail, as for the current benign image $\boldsymbol{x}$, we first accumulate the gradient momentum on $\boldsymbol{x}$ from all previous epochs and then use it to generate the adversarial initialization for the FGSM-based adversarial perturbation. The adversarial perturbation can be calculated as:
%  \begin{align}
% &\boldsymbol{G} =\operatorname{sign}\left(\nabla_{\boldsymbol{x}} \mathcal{L}(f_{ \boldsymbol{w}}(\boldsymbol{x}+\mathbf{\boldsymbol{\eta}}_{E_{t-1}} ), y)\right), \\
% &\boldsymbol{G}_{E_{t}} =\mu \cdot \boldsymbol{G}_{E_{t-1}} + \boldsymbol{G}, \\
% &\boldsymbol{\boldsymbol{\delta}}_{E_{t}}=\Pi_{[-\epsilon, \epsilon]}\left[\boldsymbol{\boldsymbol{\eta}}_{E_{t-1}}+\alpha \cdot \boldsymbol{G} \right], \label{eq:FSGM_MEP_1}\\
% &\mathbf{\boldsymbol{\eta}}_{E_{t}} =\prod_{[-\epsilon, \epsilon]}\left[\boldsymbol{\boldsymbol{\eta}}_{E_{t-1}}+\alpha \cdot \operatorname{sign}(\boldsymbol{G}_{E_{t}}) \right], \label{eq:FSGM_MEP_2}
% \end{align}
\begin{align}
    &\boldsymbol{G} &=&\ \ \ \ \ \operatorname{sign}\left(\nabla_{\boldsymbol{x}} \mathcal{L}(f_{ \boldsymbol{w}}(\boldsymbol{x}+\mathbf{\boldsymbol{\eta}}_{E_{t-1}} ), y)\right), \\[5pt]
    &\boldsymbol{G}_{E_{t}} &=&\ \ \ \ \ \mu \cdot \boldsymbol{G}_{E_{t-1}} + \boldsymbol{G}, \\[4pt]
    &\boldsymbol{\boldsymbol{\delta}}_{E_{t}} &=&\ \ \ \ \ \prod_{[-\epsilon, \epsilon]}\left[\boldsymbol{\boldsymbol{\eta}}_{E_{t-1}}+\alpha \cdot \boldsymbol{G} \right], \label{eq:FSGM_MEP_1}\\[3pt]
    &\mathbf{\boldsymbol{\eta}}_{E_{t}} &=&\ \ \ \ \ \prod_{[-\epsilon, \epsilon]}\left[\boldsymbol{\boldsymbol{\eta}}_{E_{t-1}}+\alpha \cdot \operatorname{sign}(\boldsymbol{G}_{E_{t}}) \right], \label{eq:FSGM_MEP_2}
\end{align}
where $\boldsymbol{G}$ represents the gradient of the trained model on the input image $\boldsymbol{x}$, $\boldsymbol{G}_{E_{t}}$ represents the sum of cumulative gradients $\boldsymbol{G}$ from the previous $t$ epochs, $\boldsymbol{\delta}_{E_{t}}$ represents the generated adversarial perturbation in the $t-$th epoch and $\boldsymbol{\eta}_{E_{t}}$ represents the sample adversarial initialization in the $t-$th epoch. Compared with previous sample initialization strategies, the proposed FGSM-MEP  makes more full use of prior-guided information, \emph{i.e.,} the historical gradient of the sample during the training process.

 \begin{figure}[t]
        \centering
        \includegraphics[width=0.8\linewidth]{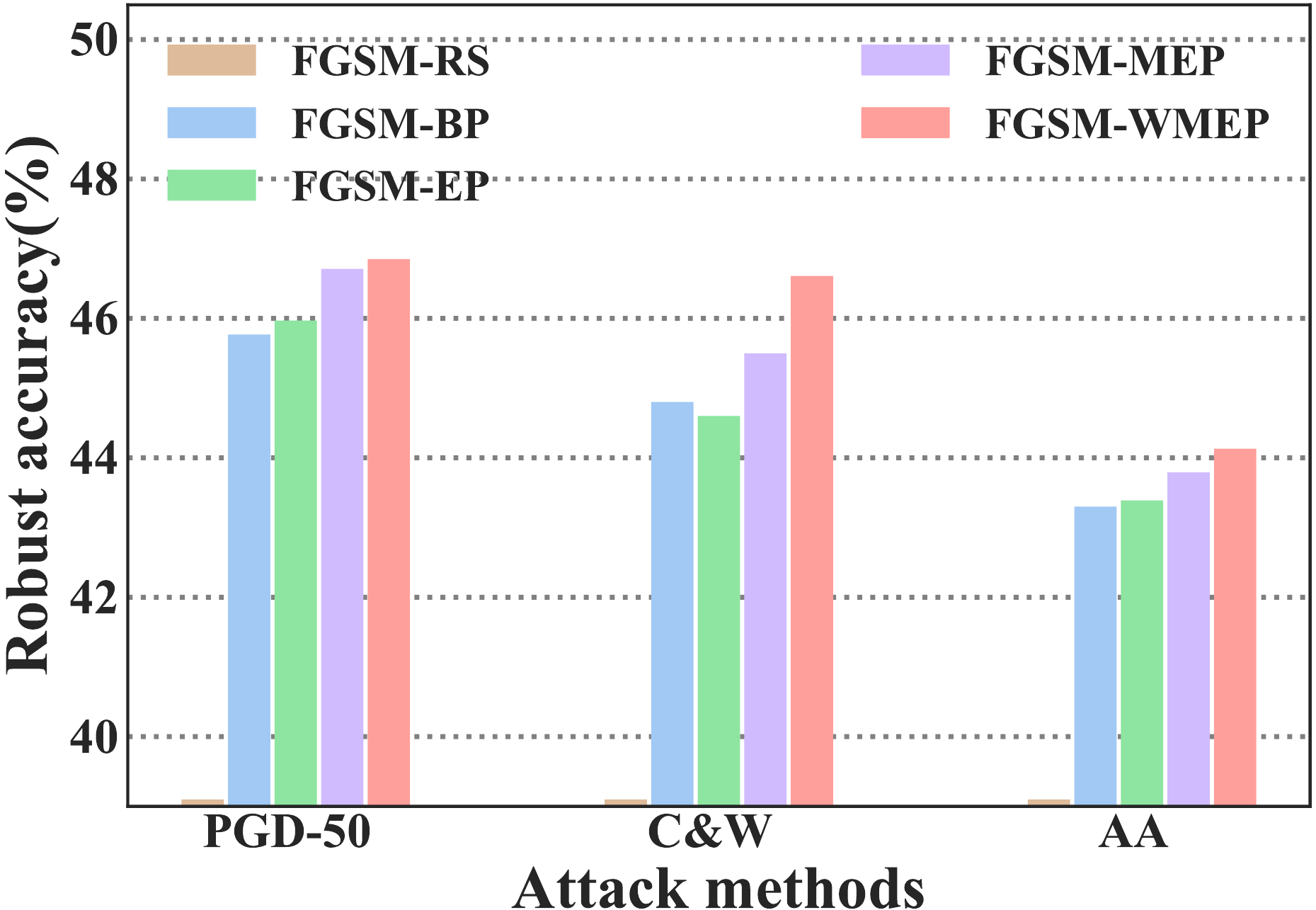}
        \label{fig:Ablation_Study}
  \caption{ Robust accuracy  of FGSM-AT with different sample initialization strategies using ResNet18 on the CIFAR-10 database. X-axis represents the attack method. Y-axis represents robust accuracy.}
\label{fig:Comparsion_Init}
\end{figure}

\subsection{The Proposed Initialization Strategy}
\label{Proposed_Initialization} 
Although the proposed FGSM-MEP can prevent catastrophic overfitting, it accumulates the negative prior-guided gradient information, resulting in limited model robustness improvement.
To overcome the shortcoming, we propose to assign different weights to different gradient momentum information of one sample for accumulation according to the gradient momentum quality. Specifically, we propose a simple and effective metric method based on the attack rate to assess the quality of the gradient momentum. The proposed evaluation metric can be defined as: 
\begin{equation}
\Gamma=1-\frac{\operatorname{Acc}(f_{\boldsymbol{w}}(\boldsymbol{x}+\boldsymbol{\delta}), y)}{\operatorname{Acc}(f_{\boldsymbol{w}}(\mathbf{x}), y)},
\end{equation}
where $\operatorname{Acc}(\cdot)$ represents the sample accuracy of the model. The higher the $\Gamma$, the higher the attack success rate of the adversarial samples generated by the current gradient, that is, the better the quality of the current gradient. We propose to accumulate the gradient momentum information with different gradient momentum weights $\Gamma$ to generate the sample initialization for training, called FGSM-WMEP. The adversarial perturbation can be calculated as:
 \begin{align}
&\boldsymbol{G} &=&\ \ \ \ \operatorname{sign}\left(\nabla_{\boldsymbol{x}} \mathcal{L}(f_{ \boldsymbol{w}}(\boldsymbol{x}+\mathbf{\boldsymbol{\eta}}_{E_{t-1}} ), y)\right), \\[5pt]
&\boldsymbol{G}_{E_{t}} &=&\ \ \ \ \mu \cdot \boldsymbol{G}_{E_{t-1}} + \Gamma \cdot \boldsymbol{G}, \\[4pt]
&\boldsymbol{\boldsymbol{\delta}}_{E_{t}}&=&\ \ \ \ \prod_{[-\epsilon, \epsilon]}\left[\boldsymbol{\boldsymbol{\eta}}_{E_{t-1}}+\alpha \cdot \boldsymbol{G} \right], \label{eq:FSGM_WMEP_1}\\[3pt]
&\mathbf{\boldsymbol{\eta}}_{E_{t}} &=&\ \ \ \ \prod_{[-\epsilon, \epsilon]}\left[\boldsymbol{\boldsymbol{\eta}}_{E_{t-1}}+\alpha \cdot \operatorname{sign}(\boldsymbol{G}_{E_{t}}) \right]. \label{eq:FSGM_WMEP_2}
\end{align}
We compare different initialization strategies, \emph{i.e.,} FGSM-BP, FGSM-EP, FGSM-MEP,  FGSM-WMEP, etc.
The robust accuracy results are shown in Fig~\ref{fig:Comparsion_Init}. It can be observed that compared to FGSM-AT initialized by random initialization, and other samples initialization strategies, the proposed FGSM-WMEP achieves the best adversarial robustness performance under all attack scenarios. Note that the proposed FGSM-WMEP does not require additional training costs compared with other sample initialization strategies. 

% it can be observed that the compared to FGSM-AT initialized by random initialization, other samples initialization strategies achieve better model robustness, and the proposed FGSM-WMEP works best. The proposed FGSM-WMEP stands out amongst them. 

\subsection{The Proposed Regularization Method}
\label{Proposed_Regularization}
As the regularization methods are widely used in fast adversarial training to improve adversarial robustness, we propose a simple and effective regularization method according to the characteristic of the proposed prior-guided initialization to further improve the adversarial robustness. In detail, as for a given input image $\boldsymbol{x}$, we initialize it with the prior-guided adversarial perturbation to obtain the perturbed image $\boldsymbol{x}+\boldsymbol{\delta}_{pgi}$. Then we adopt FGSM adversarial attack to generate the corresponding adversarial example $\boldsymbol{x}+\boldsymbol{\delta}_{a d v}$. The initialization perturbation $\boldsymbol{\delta}_{pgi}$ and FGSM-based adversarial perturbation $\boldsymbol{\delta}_{adv}$ can create the corresponding perturbed images to fool the model. However, regarding the robust model, it is expected to generate correct predictions for both types of adversarial examples. Hence, we propose a regularization term to guide the trained model in having similar correct predictions on the both types of adversarial examples. The proposed regularization term boosts the smoothness of loss function around samples by forcing the learned model to be robust to the both types of adversarial examples. Specifically, we minimize the squared $\mathcal{L}_{2}$ distance between the model predictions on the two types of adversarial examples. Our regularization term can be defined as:
\begin{equation}
    \begin{aligned}
        & & &\ \ R\left(\boldsymbol{x},\boldsymbol{\delta}_{pgi}, \boldsymbol{\delta}_{adv} ; \boldsymbol{w}\right)\\[3pt]
        & &=&\ \ \lambda\cdot\Big\|f_{\boldsymbol{w}}\left(\boldsymbol{x}+\boldsymbol{\delta}_{a d v}\right)-f_{\boldsymbol{w}}\left(\boldsymbol{x}+\boldsymbol{\delta}_{pgi}\right)\Big\|_2^2,
    \end{aligned}
\end{equation}
where $\lambda$ is the hyper-parameter. Unlike previous regularization methods ~\cite{sriramanan2020guided,sriramanan2021towards} which adopt the regularization in min-max optimization, the proposed regularization method only works in minimization optimization which requires fewer training costs. 
\par Incorporating the two proposed initialization and regularization methods, our fast adversarial training framework could be established.
Our fast adversarial training framework can be formulated as:
\begin{equation}
    \begin{aligned}
        & &\underset{\boldsymbol{w}}{\textbf{\textit{min}}}&\ \ \mathbb{E}_{(\boldsymbol{x}, y)  \sim \mathcal{D}} \Big[\mathcal{L}\left(f_{\boldsymbol{w}}\left(\boldsymbol{x}+\boldsymbol{\delta}_{adv}\right), y\right)\\
        & & &\ \ \phantom{ \mathbb{E}_{(\boldsymbol{x}, y)  \sim \mathcal{D}} \Big[\mathcal{L}}\ \ \ \ + R\left(\boldsymbol{x},\boldsymbol{\delta}_{pgi}, \boldsymbol{\delta}_{adv} ; \boldsymbol{w}\right)\Big],\\[5pt]
        & &\textbf{\textit{s.t.}}&\ \ \boldsymbol{\delta}_{adv}=\underset{\boldsymbol{\delta} \in \Omega}{\textbf{\textit{arg max}}}\ \Big[\mathcal{L}(f_{\boldsymbol{w}}(\boldsymbol{x}+\boldsymbol{\delta}), y)\Big] \\
    \end{aligned}
\end{equation}
Experiment evaluations on the cross-over of plenty of datasets demonstrate that the proposed fast adversarial training framework not only could prevent catastrophic overfitting but also could significantly enhance the adversarial robustness against adversarial examples.

\begin{algorithm}[t]
\caption{FGSM-PGK}
\label{alg:FGSM_PGK}
\begin{algorithmic}[1]
\REQUIRE
The epoch $N$, the maximal perturbation $\epsilon$, the step size $\alpha$, the dataset $\mathcal{D}$ including the benign sample $\boldsymbol{x}$ and the label ${y}$, the dataset size $M$, the network $f_{\boldsymbol{w}}(\cdot)$ with parameters $\boldsymbol{w}$, the decay factor $\mu$ for the gradient, the decay factor $\kappa$ for model weight , the hyper-parameters $\lambda$ and $\nu$, the adversarial initialization set $\mathcal{D}^{\boldsymbol{\boldsymbol{\delta}}}$ and the historical model gradient $\boldsymbol{G}^{m}$.
%The watermark image $W$, the host image $H$, the well-trained classifier $f$

\FOR{$n=1,...,N$}
\FOR{$i=1,...,M$}
\IF{$n=1$}
\STATE $\boldsymbol{\delta}_{pgi} =\boldsymbol{U}(-\epsilon ,\epsilon) $
\STATE $\boldsymbol{G} =\operatorname{sign}\left(\nabla_{\boldsymbol{x}_i} \mathcal{L}(f_{\boldsymbol{w}}(\boldsymbol{x}_i+\boldsymbol{\delta}_{pgi} ), {y}_i)\right)$
\STATE $\boldsymbol{G}^{m}_{i}= \boldsymbol{G}$
\STATE $ \boldsymbol{\boldsymbol{\delta}_{adv}} =\Pi_{[-\epsilon, \epsilon]} [ \boldsymbol{\delta}_{pgi}+ \alpha \cdot \boldsymbol{G} ]$
\STATE $\mathcal{D}^{\boldsymbol{\delta}}_{i}= \boldsymbol{\delta}_{adv}$
\STATE $ \boldsymbol{w} \leftarrow \boldsymbol{w} -\boldsymbol{\nabla}_{\boldsymbol{w}}[\mathcal{L}(f_{\boldsymbol{w}}(\boldsymbol{x}_{i} + \boldsymbol{\delta}_{adv}), {y}_i) +
\lambda \cdot\left\|f_{\boldsymbol{w}}(\boldsymbol{x}_{i}+\boldsymbol{\delta}_{adv})-{f}_{\boldsymbol{w}}(\boldsymbol{x}_{i}+\boldsymbol{\delta}_{pgi})\right\|_{2}^{2}] $
\STATE $\Lambda=\operatorname{Acc}(f_{\boldsymbol{w}}(\boldsymbol{x}_{i}+\boldsymbol{\delta}), {y}_i) / \operatorname{Acc}(f_{\boldsymbol{w}}(\boldsymbol{x}_{i}), {y}_i)$
\IF{$\Lambda<\nu$}
\STATE $\tilde{\kappa}=\kappa$
\ELSE
\STATE $\tilde{\kappa}=\Lambda/ \nu \cdot\kappa$
\ENDIF
\STATE $\tilde{\boldsymbol{w}} = \tilde{\kappa} \cdot \tilde{\boldsymbol{w}}+ (1-\tilde{\kappa}) \cdot \boldsymbol{w}$
\ELSE
\STATE $\boldsymbol{\delta}_{pgi} =\mathcal{D}^{\boldsymbol{\delta}}_{i} $
\STATE $\boldsymbol{G} =\operatorname{sign}\left(\nabla_{\boldsymbol{x}_i} \mathcal{L}(f_{\boldsymbol{w}}(\boldsymbol{x}_i+\boldsymbol{\delta}_{pgi}), {y}_i)\right)$
\STATE $\Gamma=1-{\operatorname{Acc}(f_{\boldsymbol{w}}(\boldsymbol{x}_{i}+\boldsymbol{\delta}), {y}_i)} / {\operatorname{Acc}(f_{\boldsymbol{w}}(\boldsymbol{x}_{i}), {y}_i)}$
\STATE $\boldsymbol{G}^{m}_{i}= \mu \cdot \mathcal{D}^{m}_{i} + \Gamma \cdot \boldsymbol{G}$
\STATE $ \boldsymbol{\delta}_{adv} =\Pi_{[-\epsilon, \epsilon]} [ \boldsymbol{\delta}_{pgi}+ \alpha  \cdot \boldsymbol{G} ]$
\STATE $\mathcal{D}^{\boldsymbol{\delta}}_{i} =\Pi_{[-\epsilon, \epsilon]} [ \boldsymbol{\delta}_{pgi}+ \alpha  \cdot  \operatorname{sign} (\boldsymbol{G}^{m}_{i}) ]$
\STATE $ \boldsymbol{w} \leftarrow \boldsymbol{w} -\boldsymbol{\nabla}_{\boldsymbol{w}}[\mathcal{L}(f_{\boldsymbol{w}}(\boldsymbol{x}_{i} + \boldsymbol{\boldsymbol{\delta}}_{adv}), {y}_i) +
\lambda \cdot\left\|f_{\boldsymbol{w}}(\boldsymbol{x}_{i}+\boldsymbol{\delta}_{adv})-{f}_{\boldsymbol{w}}(\boldsymbol{x}_{i}+\boldsymbol{\delta}_{pgi})\right\|_{2}^{2}] $
\STATE $\Lambda=\operatorname{Acc}(f_{\boldsymbol{w}}(\boldsymbol{x}_{i}+\boldsymbol{\delta}), {y}_i) / \operatorname{Acc}(f_{\boldsymbol{w}}(\boldsymbol{x}_{i}), {y}_i)$
\IF{$\Lambda<\nu$}
\STATE $\tilde{\kappa}=\kappa$
\ELSE
\STATE $\tilde{\kappa}=\Lambda/ \nu \cdot\kappa$
\ENDIF
\STATE $\tilde{\boldsymbol{w}} = \tilde{\kappa} \cdot \tilde{\boldsymbol{w}}+ (1-\tilde{\kappa}) \cdot \boldsymbol{w}$
\ENDIF
\ENDFOR
\ENDFOR
\end{algorithmic}
\end{algorithm}
\subsection{The Proposed Ensemble Weight}
\label{Proposed_weights}
Previous research~\cite{chen2020robust,gowal2020uncovering,wang2022self} has demonstrated that self-ensemble model weight methods, \emph{i.e.,} model weight averaging (WA)~\cite{izmailov2018averaging}, could significantly improve model generalization performance. In detail, at each training iteration, a WA model weight $\tilde{\boldsymbol{w}}$ can be obtained by calculating the exponential weighted moving average (EMA) of the trained model weight ${\boldsymbol{w}}$ with a decay rate $\kappa$. The calculation for the WA model weight is as follows:
\begin{equation}
\tilde{\boldsymbol{w}} = \kappa \cdot \tilde{\boldsymbol{w}}+ (1-\kappa) \cdot \boldsymbol{w},
\end{equation}
where $\kappa$ is set to 0.999, the WA model $\tilde{\boldsymbol{w}}$ often obtains better generalization on the unseen test data. Previous research~\cite{chen2020robust,wu2020adversarial} has explored the role of model averaging and found that using weight averaging can boost the flatter adversarial loss landscape to improve the model robustness. Especially, recent works~\cite{rebuffi2021fixing,hwang2021adversarial,wang2022self} have indicated that standard adversarial training methods combined with model weight averaging could further improve the adversarial robustness. It is noteworthy that model weight adopts the historical model weight information, \emph{i.e.,} the prior-guided model weight knowledge in the previous training, to obtain significant adversarial robustness improvement with almost no extra training costs. It prompts us to raise the question \textit{``Can we adopt the model weight technology for fast adversarial training to prevent catastrophic overfitting and improve the adversarial robustness?"}.  Unfortunately, applying the weight averaging directly for fast adversarial training also suffers from catastrophic overfitting and achieves limited adversarial robustness improvement. In detail, as shown in Table~\ref{table:cifar10_wa}, it can be observed that FGSM-RS combined with the original weight averaging (FGSM-RS-EMA) limitedly improves adversarial robustness and also encounters catastrophic overfitting. The original model weight average accumulates the model weights of all iterations with the fixed decay rate. However, during the later stage of fast adversarial training, almost all model weights suffer from catastrophic overfitting. Accumulating such weights could introduce catastrophic overfitting behavior into the final model causing it to suffer from overfitting. 
\begin{table}[t]
\centering
\caption{ Comparisons of clean and robust accuracy (\%) and training time (hour)
using ResNet18 on the CIFAR-10 dataset on the best and last checkpoints under $\ell_{\infty}=8 / 225$. Number in bold indicates the best. 
% Number in bold indicates the best of the fast AT methods. 
}
\label{table:cifar10_wa}
\setlength\tabcolsep{0.1cm}
\begin{tabular}{@{}c|c|c|c|c|c|c@{}}
\toprule
Method                         &      & Clean          & PGD-50         & C\&W           & AA             & Time (h)                \\ \midrule
\multirow{2}{*}{FGSM-RS}       & Best & 73.81          & 41.26          & 39.84          & 37.07          & \multirow{2}{*}{0.9} \\ \cmidrule(lr){2-6}
                               & Last & \textbf{83.82}          & 0.0            & 0.0            & 0.0            &                     \\ \midrule \midrule
\multirow{2}{*}{FGSM-RS-EMA~\cite{izmailov2018averaging}}   & Best & 78.75          & 50.55          & 48.79          & 45.15          & \multirow{2}{*}{0.9} \\ \cmidrule(lr){2-6}
                               & Last & 84.9           & 0.0            & 0.0            & 0.0            &                     \\ \midrule
\multirow{2}{*}{FGSM-RS-D-EMA (ours)} & Best & \textbf{79.44} & \textbf{51.02} & \textbf{49.29} & \textbf{46.63} & \multirow{2}{*}{0.9} \\ \cmidrule(lr){2-6}
                               & Last & 79.5  & \textbf{50.76} & \textbf{49.07} & \textbf{46.57} &                     \\ \bottomrule
\end{tabular}

\end{table}

\par To overcome the above shortcoming, we propose a simple yet effective dynamic decay rate mechanism for the weight average method according to the degree of model robustness. In detail, during the weight average update trajectory, we assign different decay rates to models with different robustness, \emph{i.e.,} the more robust model parameters are assigned smaller decay rates. In this way, it can reduce the impact of less robust model parameters on the final model. As we discussed in Section~\ref{Rethinking_CO}, in fast adversarial training, the quality of generated adversarial examples is directly related to the robustness of the model. 
In order not to add extra training costs, we adopt the generated adversarial example quality used for training to measure the robustness of the current training model parameters indirectly. We adopt the ratio of the accuracy of the generated adversarial examples to the accuracy of the clean examples to evaluate the adversarial example quality. It can be formulated as:
\begin{equation}
\Lambda=\frac{\operatorname{Acc}(f_{\boldsymbol{w}}(\boldsymbol{x}+\boldsymbol{\delta}), y)}{\operatorname{Acc}(f_{\boldsymbol{w}}(\mathbf{x}), y)}.
\end{equation}
Then we adopt the $\Lambda$ to obtain the dynamic decay rates for the model weight average. It can be calculated as:
\begin{equation}
\tilde{\kappa}=\left\{
	\begin{aligned}
	& \kappa \quad \Lambda < \nu, \\
	& \Lambda/\nu \cdot \kappa \quad \Lambda \geq \nu,\\
	\end{aligned}
	\right.
\end{equation}
where $\nu$ is the threshold hyper-parameter which is used to generate the dynamic decay rate $\tilde{\kappa}$. 
We replace the fixed decay rate $\kappa$ with the proposed dynamic decay rate $\tilde{\kappa}$ for the exponential weighted moving average, called D-EMA. $\Lambda$ represents the adversarial example quality. When $\Lambda$ is large, the generated adversarial example quality is inferior and the model weights updated on them are less robust and even could encounter catastrophic overfitting. Using the proposed method could reduce their impact on the weight average update trajectory. As shown in Table~\ref{table:cifar10_wa}, it can be observed that compared with the original FGSM-RS-EMA, the proposed FGSM-RS-D-EMA not only can prevent catastrophic overfitting but also can achieve better adversarial robustness performance under all attack scenarios. 

\par By assembling the proposed prior-guided knowledge, \emph{i.e.,} the prior-guided initialization and model weights, we conclude our fast adversarial training method, \emph{dubbed} FGSM-PGK. The proposed FGSM-PGK algorithm is presented in Algorithm~\ref{alg:FGSM_PGK}.

\subsection{Theoretical Analysis}
\label{sec:theoretical}
\begin{proposition}
   Suppose we have $\boldsymbol{\delta}_{pgi}$, which is the prior-guided adversarial initialization in either \textbf{FGSM-BP}, \textbf{FSGM-EP}, \textbf{FSGM-MEP} or \textbf{FGSM-WMEP}. Let
    $\boldsymbol{\hat{\delta}} _{adv}$ represent the current adversarial perturbation generated using FGSM with $\boldsymbol{\delta}_{pgi}$ as initialization, and $\alpha$ denote the step size of equations \eqref{eq:FSGM_BP}, \eqref{eq:FSGM_EP}, \eqref{eq:FSGM_MEP_1} and \eqref{eq:FSGM_MEP_2}. Assuming that Ω is a bounded set as following:
    % Let $\boldsymbol{\delta}_{pgi}$ be the prior-guided adversarial initialization in \textbf{FGSM-BP}, \textbf{FSGM-EP} or \textbf{FSGM-MEP}, $\boldsymbol{\hat{\delta}}_{adv}$ represents the current adversarial perturbation generated via FGSM using $\boldsymbol{\delta}_{pgi}$ as initialization, and $\alpha$ be the step size of \eqref{eq:FSGM_BP}, \eqref{eq:FSGM_EP}, \eqref{eq:FSGM_MEP_1} and \eqref{eq:FSGM_MEP_2}. If $\boldsymbol{\Omega}$ is a bounded set like
    \begin{equation}
        \boldsymbol{\Omega} = \Big\{\boldsymbol{\hat{\delta}}_{adv}\ :\ \|\boldsymbol{\hat{\delta}}_{adv}-\boldsymbol{\delta}_{pgi}\|^2_2\leq{\epsilon}^{2}\Big\},
    \end{equation}
    % and the step size $\alpha$ satisfies $\alpha\leq\epsilon$, it holds that
    Assuming that the step size  $\alpha$ satisfies $\alpha\leq\epsilon$, we can conclude that:
    \begin{equation}
        \label{eq:propo_1}
        \begin{aligned}
            & \mathbb{E}_{\boldsymbol{\hat{\delta}}_{adv}\sim\boldsymbol{\Omega}}\big[\|\boldsymbol{\hat{\delta}}_{adv}\|_2\big]&\leq&\ \ \sqrt{\ \mathbb{E}_{\boldsymbol{\hat{\delta}}_{adv}\sim\boldsymbol{\Omega}}\big[\|\boldsymbol{\hat{\delta}}_{adv}\|^2_2\big]}\\[5pt]
            & &\leq&\ \ \sqrt{\frac{1}{d}}\cdot\epsilon,
        \end{aligned}
    \end{equation}
    where $d$ represents the dimension of the feature space, and $\boldsymbol{\hat{\delta}}_{adv}$ is the adversarial perturbation generated by either \textbf{FGSM-BP}, \textbf{FSGM-EP}, \textbf{FSGM-MEP}, or \textbf{FSGM-WMEP}
    % where $\boldsymbol{\hat{\delta}}_{adv}$ is the adversarial perturbation generated by \textbf{FGSM-BP}, \textbf{FSGM-EP} or \textbf{FSGM-MEP}, and $d$ is the dimension of the feature space. 
\end{proposition}
\par The proof of this statement can be found in the \textbf{Appendices}. It is worth noting that the upper bound of our proposed method is $\sqrt{\frac{1}{d}}\cdot\epsilon$, which is smaller than the bound $\sqrt{\frac{d}{3}}\cdot\epsilon$ of FGSM-RS as provided in \cite{andriushchenko2020understanding}. Due to the norm of perturbation (gradient) can be treated as the convergence criteria for the non-convex optimization problem, the smaller expectation represents that the proposed prior-guided adversarial initialization will be converged to a local optimal faster than the random initialization with the same number of iterations.
\section{Experiments}
\label{sec:experiments}
\begin{table*}[t]
\centering
\caption{ Comparisons of clean and robust accuracy (\%) and training time (hour)
using ResNet18 on the CIFAR-10 dataset on the best and last checkpoints under $\ell_{\infty}=8 / 225$. Number in bold indicates the best. 
% Number in bold indicates the best of the fast AT methods. 
}
\label{table:cifar10}
\setlength\tabcolsep{0.35cm}
\begin{tabular}{c|c|c|c|c|c|c|c}
\toprule
Method           & \begin{tabular}[c]{@{}c@{}}Clean\\ Best/Last\end{tabular} & \begin{tabular}[c]{@{}c@{}}PGD-10\\ Best/Last\end{tabular} & \begin{tabular}[c]{@{}c@{}}PGD-20\\ Best/Last\end{tabular} & \begin{tabular}[c]{@{}c@{}}PGD-50\\  Best/Last\end{tabular} & \begin{tabular}[c]{@{}c@{}}C\&W\\ Best/Last\end{tabular} & \begin{tabular}[c]{@{}c@{}}AA\\  Best/Last\end{tabular} & Time (h) \\ \midrule
PGD-AT           & 82.65/82.32                                               & 53.39/53.76                                              & 52.52/52.83                                               & 52.27/52.6                                                  & 51.28/51.08                                              & 48.93/48.68                                            & 4.4 \\ \midrule \midrule
FGSM-RS~\cite{wong2020fast}          & 73.81/83.82                                               & 42.31/0.09                                                 & 41.55/0.04                                                 & 41.26/0.02                                                  & 39.84/0.00                                               & 37.07/0.00                                              & 0.9   \\ \midrule
FGSM-CKPT~\cite{kim2021understanding}        & \textbf{90.29/90.29}                                      & 41.96/41.96                                                & 39.84/39.84                                                & 39.15/39.15                                                 & 41.13/41.13                                              & 37.15/37.15                                             & 1.3  \\ \midrule
FGSM-SDI~\cite{jia2022boosting}         & 84.86/85.25                                               & 53.73/53.18                                                & 52.54/52.05                                                & 52.18/51.79                                                 & 51.0/50.29                                               & 48.50/47.91                                             & 1.4  \\ \midrule
NuAT~\cite{sriramanan2021towards}             & 81.58/81.38                                               & 53.96/53.52                                                & 52.9/52.65                                                 & 52.61/52.48                                                 & \textbf{51.3}/50.63                                               & 49.09/48.70                                             & 1.7 \\ \midrule
GAT~\cite{sriramanan2020guided}              & 79.79/80.41                                               & 54.18/53.29                                                & 53.55/52.06                                                & 53.42/51.76                                                 & 49.04/49.07                                              & 47.53/46.56                                             & 1.8 \\ \midrule
FGSM-GA~\cite{andriushchenko2020understanding}          & 83.96/84.43                                               & 49.23/48.67                                                & 47.57/46.66                                                & 46.89/46.08                                                 & 47.46/46.75                                              & 43.45/42.63                                             & 3.0 \\ \midrule
Free-AT (m=8)~\cite{shafahi2019adversarial}    & 80.38/80.75                                               & 47.1/45.82                                                 & 45.85/44.82                                                & 45.62/44.48                                                 & 44.42/43.73                                              & 42.17/41.17                                             & 3.6
 \\ \midrule 
FGSM-PGI~\cite{jia2022prior}        & 81.72/81.72                                               & 55.18/55.18                                                & 54.36/54.36                                                & 54.17/54.17                                                 & 50.75/50.75                                              & 49.0/49.0                                               & 1.2
\\ \midrule

FGSM-PGK (ours) &
 81.58/81.58                                               & \textbf{56.08/56.08}                                       & \textbf{55.51/55.51}                                       & \textbf{55.31/55.31}                                        & 51.17/\textbf{51.17}                                     & \textbf{49.51/49.51}                                    & 1.2  \\ \bottomrule
\end{tabular}

\end{table*}

To evaluate the proposed method, we conduct extensive experiments across various network architectures and benchmark datasets which include CIFAR-10~\cite{krizhevsky2009learning}, CIFAR-100~\cite{krizhevsky2009learning}, Tiny ImageNet~\cite{deng2009imagenet}, and ImageNet~\cite{deng2009imagenet}. We compare the proposed method with the state-of-the-art fast adversarial training methods which consist of Free-AT~\cite{shafahi2019adversarial}, FGSM-RS~\cite{wong2020fast}, FGSM-GA~\cite{andriushchenko2020understanding}, GAT~\cite{sriramanan2020guided},  FGSM-CKPT~\cite{kim2021understanding}, NuAT~\cite{sriramanan2021towards}, and  FGSM-SDI~\cite{jia2022boosting}. 
We also compare the proposed method with an advanced multi-step adversarial training method which is a PGD-based adversarial training method with an early stopping, \emph{i.e.,} PGD-AT~\cite{rice2020overfitting}.

\subsection{Detailed Experimental Settings}
\subsubsection{Image Data Sets} Four benchmark image databases are used to conduct comparison experiments including CIFAR-10~\cite{krizhevsky2009learning}, CIFAR-100~\cite{krizhevsky2009learning}, Tiny ImageNet~\cite{deng2009imagenet}, and ImageNet~\cite{deng2009imagenet}. In detail, CIFAR-10 and CIFAR-100 include 50,000 training color images and 10,000 test color images with the size $32 \times 32$. They cover 10 classes and 100 classes, respectively. ImageNet and Tiny ImageNet are large-scale image classification data sets, which are harder to achieve adversarial robustness on them. ImageNet is a large-scale hierarchical image database that covers 1000 classes. The training images are resized to $224 \times 224$ to conduct experiments. Tiny ImageNet is a subset of ImageNet. It covers 200 classes, which class includes 600 color images with the size $64 \times 64$. As for ImageNet and Tiny ImageNet, following the setting of the previous works~\cite{lee2020adversarial,wong2020fast,jia2022boosting}, we adopt the validation data sets of them to conduct the evaluation experiments due to the test data sets of them have no labels. 

\begin{table}[t]
\centering
\caption{ Comparisons of clean and robust accuracy (\%) and training time (hour)
using WideResNet34-10 on the CIFAR-10 dataset on the best checkpoints under $\ell_{\infty}=8 / 225$. Number in bold indicates the best. 
% Number in bold indicates the best of the fast AT methods. 
}
\label{table:cifar10_wide}
\setlength\tabcolsep{0.28cm}
\begin{tabular}{c|c|c|c|c}
\toprule
Method          & Clean          & PGD-50         & AA             & Time (h) \\ \midrule
PGD-AT~\cite{rice2020overfitting}          & 85.17          & 54.87          & 51.67          & 31.9     \\  \midrule \midrule
FGSM-RS~\cite{wong2020fast}        & 74.3           & 40.9           & 38.4           & 5.8      \\ \midrule
FGSM-CKPT~\cite{kim2021understanding}       & \textbf{91.84} & 42.25          & 40.46          & 8.7      \\ \midrule
FGSM-SDI~\cite{jia2022boosting}       & 86.4           & 54.6           & 51.17          & 9.4      \\ \midrule
NuAT~\cite{sriramanan2021towards}            & 85.30          & 53.75          & 50.06          & 11.8     \\ \midrule
GAT~\cite{sriramanan2020guided}             & 85.17          & 54.97          & 50.01          & 12.9     \\ \midrule
FGSM-GA~\cite{andriushchenko2020understanding}         & 82.1           & 46.9           & 45.7           & 20.3     \\ \midrule
Free-AT~\cite{shafahi2019adversarial}         & 80.1           & 46.3           & 43.9           & 23.7     \\ \midrule
FGSM-PGI~\cite{jia2022prior}  & 85.09          & 56.4           & 50.11          & 8.3      \\ \midrule
FGSM-PGK (ours) &  83.32          & \textbf{59.79} & \textbf{53.28} & 8.3      \\ \bottomrule
\end{tabular}

\end{table}

\begin{table}[t]
\centering
\caption{ Comparisons of clean and robust accuracy (\%) and training time (minute)
using ResNet18 on the CIFAR-10 dataset on the last checkpoints under $\ell_{\infty}=8 / 225$. All models are trained using a cyclic learning rate strategy. Number in bold indicates the best. 
% Number in bold indicates the best of the fast AT methods. 
}
\label{table:cifar10_cyclic}
\setlength\tabcolsep{0.07cm}
\begin{tabular}{c|c|c|c|c}
\toprule
Method          & \begin{tabular}[c]{@{}c@{}}Clean\\ Best/Last\end{tabular} & \begin{tabular}[c]{@{}c@{}}PGD-50\\ Best/Last\end{tabular} & \begin{tabular}[c]{@{}c@{}}AA\\ Best/Last\end{tabular} & Time (min) \\ \midrule
FGSM-RS~\cite{wong2020fast}         & 83.75/83.75                                               & 46.11/46.11                                                & 42.92/42.92                                            & 15         \\ \midrule
FGSM-CKPT~\cite{kim2021understanding}       & \textbf{89.08/89.08}                                      & 37.69/37.69                                                & 35.66/35.66                                            & 23         \\ \midrule
FGSM-SDI~\cite{jia2022boosting}        & 82.08/82.08                                               & 50.33/50.33                                                & 46.21/46.21                                            & 24         \\ \midrule
NuAT~\cite{sriramanan2021towards}            & 76.23/76.23                                               & 50.64/50.64                                                & 46.33/46.33                                            & 30         \\ \midrule
GAT~\cite{sriramanan2020guided}            & 81.91/81.91                                               & 49.62/49.62                                                & 45.24/46.24                                            & 33         \\ \midrule
FGSM-GA~\cite{andriushchenko2020understanding}         & 80.83/80.83                                               & 47.54/47.54                                                & 43.06/43.06                                            & 53         \\ \midrule
Free-AT~\cite{shafahi2019adversarial}         & 75.22/75.22                                               & 43.72/43.72                                                & 40.3/40.3                                              & 58         \\ \midrule
FGSM-PGI~\cite{jia2022prior}  & 80.68/80.68                                               & 51.5/51.5                                                  & 46.65/46.65                                            & 22         \\ \midrule
FGSM-PGK (ours) & 
 80.85/80.85                                               & \textbf{52.97/52.97}                                       & \textbf{46.79/46.79}                                   & 22         \\ \bottomrule
\end{tabular}

\end{table}

\begin{table*}[t]
\centering
\caption{ Comparisons of clean and robust accuracy (\%) and training time (hour)
using ResNet18 on the CIFAR-100 dataset on the best and last checkpoints under $\ell_{\infty}=8 / 225$. Number in bold indicates the best. 
% Number in bold indicates the best of the fast AT methods. 
}
\label{table:cifar100}
\setlength\tabcolsep{0.35cm}
\begin{tabular}{c|c|c|c|c|c|c|c}
\toprule
Method           & \begin{tabular}[c]{@{}c@{}}Clean\\ Best/Last\end{tabular} & \begin{tabular}[c]{@{}c@{}}PGD-10\\ Best/Last\end{tabular} & \begin{tabular}[c]{@{}c@{}}PGD-20\\ Best/Last\end{tabular} & \begin{tabular}[c]{@{}c@{}}PGD-50\\  Best/Last\end{tabular} & \begin{tabular}[c]{@{}c@{}}C\&W\\ Best/Last\end{tabular} & \begin{tabular}[c]{@{}c@{}}AA\\  Best/Last\end{tabular} & Time (h) \\ \midrule
PGD-AT~\cite{rice2020overfitting}           & 57.52/57.5                                                & 29.6/29.54                                                 & 28.99/29.0                                                 & 28.87/28.9                                                  & 28.85/27.6                                               & 25.48/25.58                                             & 4.7 \\ \midrule \midrule
FGSM-RS~\cite{wong2020fast}          & 49.85/60.55                                               & 22.47/0.45                                                 & 22.01/0.25                                                 & 21.82/0.19                                                  & 20.55/0.25                                               & 18.29/0.00                                              & 0.9   \\ \midrule
FGSM-CKPT~\cite{kim2021understanding}        & \textbf{60.93/60.93}                                      & 16.58/16.69                                                & 15.47/15.61                                                & 15.19/15.24                                                 & 16.4/16.6                                                & 14.17/14.34                                             & 1.4  \\ \midrule
FGSM-SDI~\cite{jia2022boosting}         & 60.67/60.82                                               & 31.5/30.87                                                 & 30.89/30.34                                                & 30.6/30.08                                                  & 27.15/27.3                                               & 25.23/25.19                                             & 1.7  \\ \midrule
NuAT~\cite{sriramanan2021towards}             & 59.71/59.62                                               & 27.54/27.07                                                & 23.02/22.72                                                & 20.18/20.09                                                 & 22.07/21.59                                              & 11.32/11.55                                             & 1.9  \\ \midrule
GAT~\cite{sriramanan2020guided}              & 57.01/56.07                                               & 24.55/23.92                                                & 23.8/23.18                                                 & 23.55/23.0                                                  & 22.02/21.93                                              & 19.60/19.51                                             & 2.0   \\ \midrule
FGSM-GA~\cite{andriushchenko2020understanding}          & 54.35/55.1                                                & 22.93/20.04                                                & 22.36/19.13                                                & 22.2/18.84                                                  & 21.2/18.96                                               & 18.88/16.45                                             & 3.1  \\ \midrule
Free-AT (m=8)~\cite{shafahi2019adversarial}    & 52.49/52.63                                               & 24.07/22.86                                                & 23.52/22.32                                                & 23.36/22.16                                                 & 21.66/20.68                                              & 19.47/18.57                                             & 3.8  \\ \midrule
FGSM-PGI~\cite{jia2022prior}        & 58.78/58.81                                               & 31.78/31.6                                                 & 31.26/31.06                                                & 31.14/30.88                                                 & 28.06/27.72                                              & 25.67/25.42                                             & 1.3  \\ \midrule
FGSM-PGK (ours) &  
 56.27/58.13                                              & \textbf{33.15/32.38}                                         & \textbf{32.85/31.9}                                       & \textbf{32.83/31.87}                                        & \textbf{28.39/27.95}                                     & \textbf{26.86/26.35}                                    & 1.3  \\ \bottomrule
\end{tabular}
\end{table*}
\begin{table*}[t]
\centering
\caption{ Comparisons of clean and robust accuracy (\%) and training time (hour) on the Tiny ImageNet database using PreActResNet18 with different adversarial training methods under $\ell_{\infty}=8 / 225$. Number in bold indicates the best. 
% Number in bold indicates the best of the fast AT methods. 
}
\label{table:Tiny_ImageNet}
\setlength\tabcolsep{0.35cm}
\begin{tabular}{c|c|c|c|c|c|c|c}
\toprule
Method       & \begin{tabular}[c]{@{}c@{}}Clean\\ Best/Last\end{tabular} & \begin{tabular}[c]{@{}c@{}}PGD-10\\ Best/Last\end{tabular} & \begin{tabular}[c]{@{}c@{}}PGD-20\\ Best/Last\end{tabular} & \begin{tabular}[c]{@{}c@{}}PGD-50\\ Best/Last\end{tabular} & \begin{tabular}[c]{@{}c@{}}C\&W\\ Best/Last\end{tabular} & \begin{tabular}[c]{@{}c@{}}AA\\ Best/Last\end{tabular} & Time (h) \\ \midrule
PGD-AT~\cite{rice2020overfitting}       & 43.6/45.28                                                  & 20.2/16.12                                                   & 19.9/15.6                                                    & 19.86/15.4                                                   & 17.5/14.28                                                 & 16.0/12.84                                               &   30.1\\ \midrule \midrule
FGSM-RS~\cite{wong2020fast}      & 44.98/45.18                                                 & 17.72/0.00                                                   & 17.46/0.00                                                   & 17.36/0.00                                                   & 15.84/0.00                                                 & 14.08/0.00                                               & 5.5    \\ \midrule
FGSM-CKPT~\cite{kim2021understanding}    & \textbf{49.98/49.98}                                        & 9.20/9.20                                                    & 9.20/9.20                                                    & 8.68/8.68                                                    & 9.24/9.24                                                  & 8.10/8.10                                                & 8.3   \\ \midrule
FGSM-SDI~\cite{jia2022boosting}     & 46.46/47.64                                                 & 23.22/19.84                                                  & 22.84/19.36                                                  & 22.76/19.16                                                  & 18.54/16.02                                                & 17.0/14.10                                               & 9.4   \\ \midrule
NuAT~\cite{sriramanan2021towards}         & 42.9/42.42                                                  & 15.12/13.78                                                  & 14.6/13.34                                                   & 14.44/13.2                                                   & 12.02/11.32                                                & 10.28/9.56                                               & 11   \\ \midrule
GAT~\cite{sriramanan2020guided}          & 42.16/41.84                                                 & 15.02/14.44                                                  & 14.5/13.98                                                   & 14.44/13.8                                                   & 11.78/11.48                                                & 10.26/9.74                                               & 11.1   \\ \midrule
FGSM-GA~\cite{andriushchenko2020understanding}      & 43.44/43.44                                                 & 18.86/18.86                                                  & 18.44/18.44                                                  & 18.36/18.36                                                  & 16.2/16.2                                                  & 14.28/14.28                                              & 17.6   \\ \midrule
Free-AT (m=8)~\cite{shafahi2019adversarial} & 38.9/40.06                                                  & 11.62/8.84                                                   & 11.24/8.32                                                   & 11.02/8.2                                                    & 11.00/8.08                                                 & 9.28/7.34                                                & 22.9   \\ \midrule
FGSM-PGI~\cite{jia2022prior}     & 43.32/45.88                                                 & 23.8/22.02                                                   & 23.4/21.7                                                    & 23.38/21.6                                                    & 19.28/17.44                                                & 17.56/15.50                                              & 8.6   \\ \midrule

FGSM-PGK (ours)    &  43.57/43.35                                                 & \textbf{25.48/24.8}                                         & \textbf{25.35/24.64}                                         & \textbf{25.26/24.51}                                         & \textbf{19.99/19.44}                                       & \textbf{18.33/17.86}                                   & 8.6 \\ \bottomrule
\end{tabular}
\end{table*}
\par 
\subsubsection{Experimental Setups}  As for CIFAR-10, we adopt the ResNet18~\cite{he2016deep} 
and WideResNet34-10~\cite{zagoruyko2016wide} which are wide to evaluate adversarial robustness to conduct evaluation experiments. As for CIFAR-100, we adopt the ResNet18~\cite{he2016deep}  to conduct evaluation experiments. As for Tiny ImageNet, we adopt the  PreActResNet18~\cite{he2016identity}  to conduct evaluation experiments. As for ImageNet, we adopt the ResNet50~\cite{he2016deep} to conduct evaluation experiments. 
On CIFAR-100, CIFAR-100, and Tiny ImageNet, following the previous works~\cite{rice2020overfitting,jia2022boosting}, the total training epoch is set to 110. The initial learning rate is set to 0.1 and we adopt a factor of 0.1 to decay the learning rate  at the 100-th and 105-th epoch. The SGD~\cite{qian1999momentum} momentum optimizer with the weight decay of $5 \times 10^{-4}$ is used to conduct evaluation experiments. On ImageNet,  following the previous works~\cite{shafahi2019adversarial,wong2020fast,jia2022boosting},  the total training epoch is set to 90. The initial learning rate is set to 0.1 and we adopt a factor of 0.1 to decay the learning rate  at the 30-th and 60-th epoch. The SGD~\cite{qian1999momentum} momentum optimizer with the weight decay of $5 \times 10^{-4}$ is used to conduct evaluation experiments. To evaluate the model adversarial robustness, a series of adversarial attack methods which consist of FGSM~\cite{goodfellow2014explaining}, PGD attack of 10-steps (PGD-10)~\cite{madry2017towards},  PGD attack of 20-steps (PGD-20)~\cite{madry2017towards},  PGD attack of 50-steps (PGD-50)~\cite{madry2017towards},  C\&W~\cite{carlini2017towards}, and AA~\cite{croce2020reliable}, are used to perform the evaluation.
Following the training settings of previous works~\cite{wong2020fast,kim2021understanding,sriramanan2021towards}, the maximal perturbation $\epsilon$ is set to $8/255$ under the $\ell_{\infty}$ on CIFAR-10, CIFAR-100, and Tiny ImageNet. And the maximal perturbation $\epsilon$ is set to $8/255$ under the $\ell_{\infty}$ on ImageNet. We adopt the clean accuracy on clean images and robust accuracy on adversarial examples as the evaluation metrics to perform experiments.  All evaluation experiments are conducted on a single NVIDIA Tesla V100 to count the training times. A cyclical learning rate~\cite{smith2017cyclical} is also performed to conduct evaluation experiments. Note that we not only report the experiment results of the last checkpoint but also report the experiment results of the checkpoint with the best accuracy under PGD-10. 

\subsection{Comparison Experiments}
\label{Comparison_Experiments}
\subsubsection{Results on CIFAR-10.}  
\par We use ResNet18 as the backbone. The comparison experiment results on CIFAR-10 are shown in Table~\ref{table:cifar10}. It can be observed that the proposed FGSM-PGK not only prevents catastrophic overfitting but also achieves excellent adversarial robustness performance improvement. Compared with state-of-the-art multi-step PGD-AT~\cite{rice2020overfitting}, the proposed FGSM-PGK can obtain better adversarial robustness under all attack scenarios with less training cost. 
In particular, under the strongest attack method (AA), PGD-AT achieves the performance of about 48.7\%, but the proposed FGSM-PGK achieves the performance of about 49.5\%, which indicates that single-step adversarial training with prior-guided knowledge has the potential to outperform multi-step adversarial training. More importantly, our FGSM-PGK is about 3.6 times faster than PGD-AT. Compared with previous fast adversarial training methods, our FGSM-PGK obtains the best adversarial robustness under all attack scenarios on the best and last
checkpoints. For example, under the PGD-50 attack, the most robust one among the previous fast adversarial training methods achieves an accuracy of about 53\%, but our FGSM-PGK achieves an accuracy of about 55\%. 
In terms of training efficiency, our proposed FGSM-PGK only takes 1.2 hours to train, which is faster than previous fast adversarial training methods. Specifically, the entire training process of FGSM-PGK is 1.1 times faster than FGSM-CKPT, 1.2 times faster than FGSM-SDI, 1.4 times faster than NuAT, 1.5 times faster than GAT, 2.5 times faster than FGSM-GA, and 3 times faster than Free-AT. These results demonstrate the superior efficiency of our proposed method compared to other state-of-the-art techniques.
% In terms of training efficiency, the proposed FGSM-PGK only requires 1.2 hours, which is faster than the previous fast adversarial training variants. The whole training process of FGSM-PGK is about 1.1 times faster than FGSM-CKPT, 1.2 times faster than FGSM-SDI, 1.4 times faster than NuAT, 1.5 times faster than GAT, 2.5 times faster than FGSM-GA, and 3 times faster than Free-AT. 
% We also adopt a cyclic learning rate strategy~\cite{smith2017cyclical} to conduct comparative experiments by using ResNet18.The results are shown in Fig.~\ref{fig:comparsion_cyclic}. It can be observed that our FGSM-PGK achieves the best adversarial robustness under
% all attack scenarios compared with previous fast adversarial training methods. It demonstrates using the prior-guided knowledge can significantly improve the adversarial robustness.  

\begin{table}[t]
\centering
\caption{ Comparisons of clean and robust accuracy (\%) and training time (hour) on the ImageNet database using ResNet50 with different adversarial training methods under $\ell_{\infty}=4 / 225$. Number in bold indicates the best. 
% Number in bold indicates the best of the fast AT methods. 
}
\label{table:ImageNet}
\setlength\tabcolsep{0.21cm}
\begin{tabular}{c|c|c|c|c}
\toprule
Method        & Clean          & PGD-10         & PGD-50        & \begin{tabular}[c]{@{}c@{}} Time (h)\end{tabular} \\ \midrule
PGD-AT~\cite{rice2020overfitting}        & 59.19          & 35.87          & 35.41         & 211.2                                                         \\ \midrule \midrule
Free-AT (m=4)~\cite{shafahi2019adversarial} & 63.42          & 33.22          & 33.08         & 127.7                                                         \\ \midrule
FGSM-RS~\cite{wong2020fast}       & 63.65          & 35.01          & 32.66         & 44.5                                                          \\ \midrule
FGSM-PGI~\cite{jia2022prior}   & 64.32          & 36.24          & 34.93         & 63.7                                                          \\ \midrule
  FGSM=PGK (ours)     & 
 \textbf{66.24} & \textbf{37.13} & \textbf{35.7} & 63.7                                                          \\ \bottomrule
\end{tabular}
\end{table}
\begin{figure}[htbp]
\begin{center}
   \includegraphics[width=0.95\linewidth]{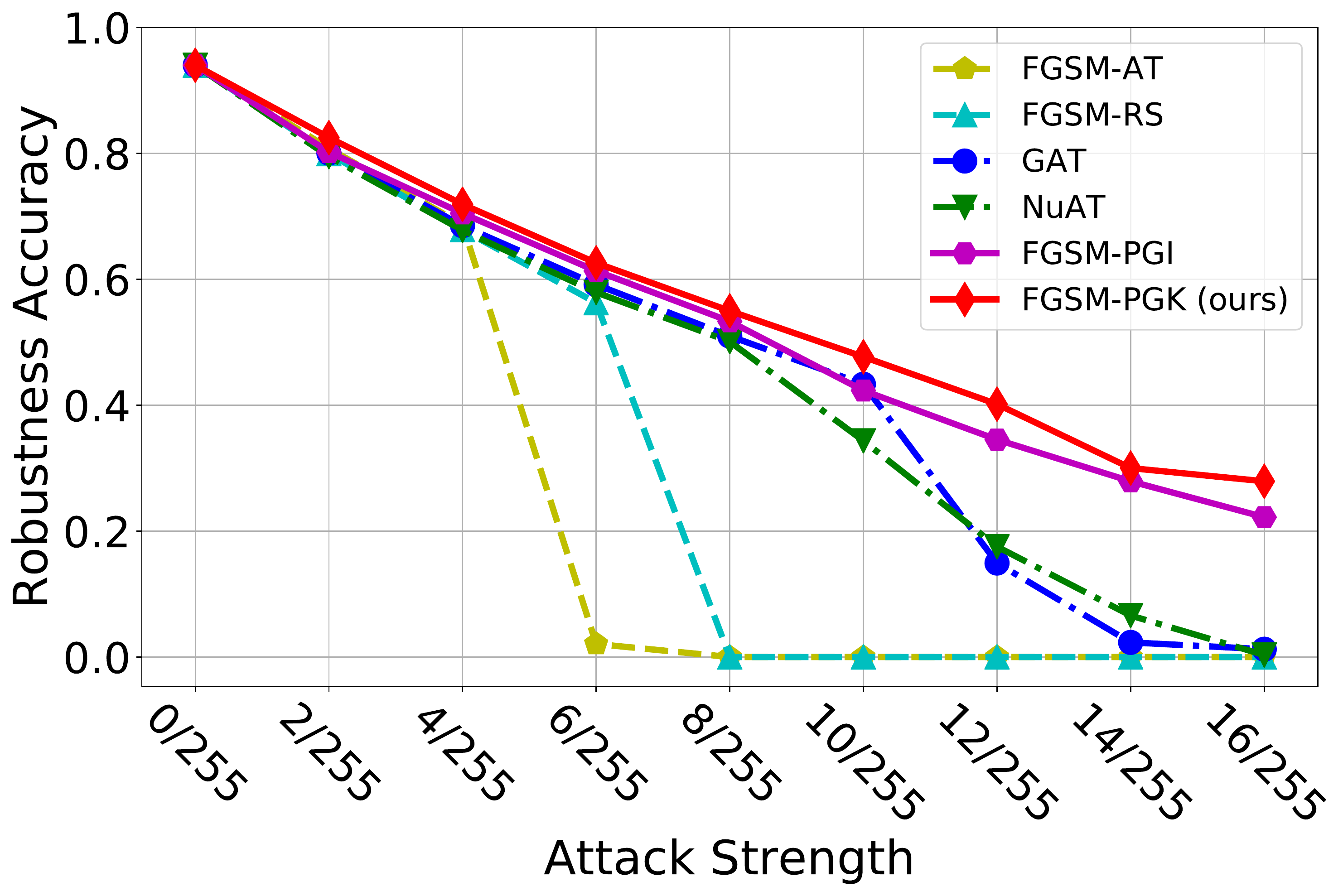}
\end{center}
\vspace{-4mm} 
\caption{ Adversarial robustness accuracy of different fast adversarial training methods under PGD-50 with different attack strengths ($\ell_{\infty}= 0/225 \rightarrow 16/255$) on CIFAR-10.}
% robustness performance of different advanced fast adversarial training methods for CIFAR-10 with the same training setting and evaluation attack strengths under PGD-50.}
% The target network provides feedbacks to update the strategy network. }
\label{fig:attack_strength}
\vspace{-4mm} 
\end{figure}
\begin{figure*}[htbp]
\begin{center}
   \includegraphics[width=1\linewidth]{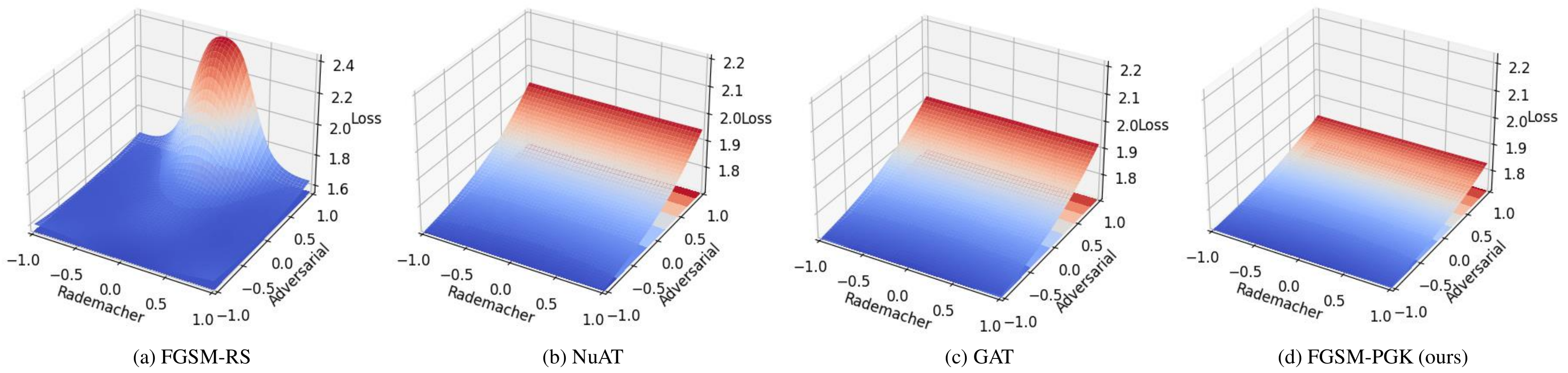}
\end{center}
\vspace{-5mm}
\caption{We visualize the loss landscape of different fast adversarial training methods, \emph{i.e.,} FGSM-RS~\cite{wong2020fast}, NuAT~\cite{sriramanan2021towards}, GAT~\cite{sriramanan2020guided}, and the proposed FGSM-PGK. 1000 images randomly sampled from CIFAR-10 testing dataset are used to plot their cross-entropy loss as it varied in two directions: a Rademacher (random) direction denoted as $v$, and an adversarial direction denoted as $u$. The Rademacher direction is calculated as $v \sim \operatorname{Rademacher}(\eta)$, where $\eta$ is set to 8/255. The adversarial direction is calculated as $u=\eta \operatorname{sign}(\nabla_{x} f(\hat{x}))$. It's worth noting that we used the same adversarial attack (PGD-50) to find the adversarial direction.}

% \caption{Visualization of the loss landscape on CIFAR10 of different regularization methods, \emph{i.e.,} the baseline FGSM-BR, FGSM-BR with the guided regularization(min-max),  FGSM-BR with the nuclear-norm regularization(min-max) and FGSM-BR with our proposed regularization. We randomly select 1000 test images for CIFAR-10 testing dataset and plot the cross-entropy loss along a spatial variation composed of two directions, \emph{i.e.,} a Rademacher (random) direction $r_{1}$ and an adversarial direction $r_{2}$. The Rademacher (random) direction is calculated as:  $r_{1} \sim \operatorname{Rademacher}(\eta)$, where $\eta$ is $8/255$. The adversarial direction is calculated as: $r_{2}=\eta \operatorname{sign}(\nabla_{x} f(\mathbf{x}; \boldsymbol{\theta}))$. Note the same adversarial attack  (PGD-50) are conducted on the same images for the visualization.}
% The strategy network is trained using REINFORCE gradients. 
%some non-differentiable operations (\emph{e.g.} choosing the iteration times) related to attack break gradient flow from the target network to the strategy network. As an alternative approach, REINFORCE algorithm~\cite{williams1992simple} is applied to optimize the strategy network and we utilize the so-called ``REINFORCE gradient" to update the strategy network.

% The target network provides feedbacks to update the strategy network. }
\label{fig:Fig_loss_landscape}
\vspace{-5mm} 
\end{figure*}

\par When using WideResNet34-10 as the backbone, our experimental results, presented in Table~\ref{table:cifar10_wide}, show that our proposed FGSM-PGK achieves the best adversarial robustness compared to previous fast adversarial training methods and the advanced PGD-AT, under all attack scenarios.
% As for WideResNet34-10 as the backbone, the comparison experiment results are shown in Table~\ref{table:cifar10_wide}. 
% It can be observed that compared with previous fast adversarial training variants and the advanced PGD-AT, the proposed FGSM-PGK achieves the best adversarial robustness under all attack scenarios. 
Particularly, under AA attack, the most robust one among fast adversarial training variants obtains a performance of about 50\% which is below the performance of about 51\% achieved by PGD-AT. It is because the structure of WideResNet34-10 is more complicated and non-linear than that of ResNet18, which makes it hard for fast adversarial training methods to generate high-quality adversarial examples for training. But our FGSM-PGK obtains the performance of about 53\%. Besides, PGD-AT requires about 31.9 hours for training,
while our FGSM-PKG only requires about 8.3 hours for
training.

\par Moreover, we adopt ResNet18 with a cyclic
learning rate strategy~\cite{smith2017cyclical} to conduct comparative experiments using  on CIFAR-10. Following the default training settings of their original paper, we set the maximum learning rate of the proposed method to 0.2. All the models are trained for 30 epochs. The results are shown in Table~\ref{table:cifar10_cyclic}. The results of our experiments show that the proposed FGSM-PGK method exhibits similar characteristics to the models trained using a multi-step learning rate strategy. In particular, among other FAT methods, our method achieves the best adversarial robustness across all attack scenarios. In detail, compared to the FGSM-GA, our proposed FGSM-PGK achieves higher robustness accuracy, for example, approximately 46.79\% under AA attack, while FGSM-GA achieves only 43.06\% accuracy. Furthermore, our training process is considerably more efficient than other state-of-the-art AT methods, with our method being approximately 3 times faster than Free-AT, 2.5 times faster than FGSM-GA, and 1.4 times faster than GAT and NuAT.  It demonstrates that using prior-guided knowledge can significantly improve adversarial robustness.

\subsubsection{Results on CIFAR-100.}  
\par For CIFAR-100, we use ResNet18 as the backbone. The comparison experiment results on CIFAR-100 are shown in Table~\ref{table:cifar100}. Similar experiments result as on
CIFAR-10 can be observed, \emph{i.e.,} the proposed FGSM-PGK achieves the best adversarial robustness compared with the previous fast adversarial training methods. In detail, compared with the best robust method among fast adversarial training methods which achieves the performance of about 22\% under C\&W attack, the proposed method achieves the performance of about 28\% under C\&W attack. It demonstrates that fast adversarial training combined with prior-guided knowledge could effectively improve adversarial robustness. From the perspective of training efficiency,  we can observe similar results as on CIFAR-10, \emph{i.e.,} the proposed FGSM-PGK requires less training cost than the majority of fast adversarial training methods. 
Though the proposed FGSM-PKG needs a little more training time than FGSM-RS, it not only prevents catastrophic overfitting but also achieves advanced adversarial robustness, \emph{i.e.,} on the last checkpoint, it also obtains comparable robustness to the model with the best checkpoint. Surprisingly, even compared with the advanced PGD-AT, our FGSM-PGK improves the accuracy by about 1\% under the strong AA attack. Besides, the proposed FGSM-PGK is about 3.6 times faster than PGD-AT, \emph{i.e.,} PGD-AT requires 4.7 hours for training but our FGSM-PGK only requires 1.3 hours.

\subsubsection{Results on Tiny ImageNet.} 
\par As for Tiny ImageNet, we adopt PreActResNet18 as the backbone, the comparison experiment results on Tiny ImageNet are shown in Table~\ref{table:Tiny_ImageNet}. The Tiny ImageNet database covers more images and classes than CIFAR-10 and CIFAR-100, which requires more training costs to perform adversarial training. In detail, the multi-step PGD-AT requires about 30.1 hours to achieve adversarial robustness. But the proposed FGSM-PGK only requires about 8.6 hours which is 3.6 times faster than PGD-AT and obtains better adversarial robustness than PGD-AT under all attack scenarios. For example, PGD-AT achieves an accuracy of about 17\% and 14\% with the best and last checkpoints under C\&W attack. But our FGSM-PGK obtains the accuracy of about 20\% and 19\%  with the best and last checkpoints under C\&W attack. Besides, our FGSM-PGK also achieves the best adversarial robustness compared with state-of-the-art fast adversarial training methods under all attack scenarios. Specifically, under the AA attack, the most robust one among previous fast adversarial training methods obtains an accuracy of about 17\% and 14\%  with the best and last checkpoints, respectively. But our FGSM-PGK obtains the accuracy of about 18\% and 17\%  with the best and last checkpoints, respectively. From the perspective of training efficiency, similar efficiency
comparison results as on CIFAR-10 can be observed. 

\subsubsection{Results on ImageNet.} 
\par We adopt ResNet50 as the backbone for conducting comparative experiments on ImageNet. The maximum perturbation strength $\epsilon$  is set to 4/255, following the training settings of~\cite{shafahi2019adversarial,wong2020fast}. We compare our proposed FGSM-PGK method with several efficient FAT methods, including  Free-AT~\cite{shafahi2019adversarial}, FGSM-RS~\cite{wong2020fast}, and FGSM-PGI~\cite{jia2022prior} and one advanced standard adversarial training method, \emph{i.e.,} PGD-AT~\cite{rice2020overfitting}. The results are presented in Table~\ref{table:ImageNet}. Our proposed FGSM-PGK method achieves higher clean and robust accuracy than PGD-AT. Moreover, it significantly outperforms Free-AT, FGSM-RS, and FGSM-PGI in terms of both clean and robust accuracy. In addition, our FGSM-PGK is 3.3 times faster than the advanced PGD-AT in terms of training efficiency. Similar observations are made for other databases as well.

% As for ImageNet, We conduct comparative experiments using ResNet50 as the backbone. Following the training settings of~\cite{shafahi2019adversarial,wong2020fast}, the maximum perturbation strength $\epsilon$ is set to 4/255. We compare our proposed FGSM-PGK with several efficient FAT methods, \emph{i.e.,} Free-AT~\cite{shafahi2019adversarial}, FGSM-RS~\cite{wong2020fast}, PGD-AT~\cite{rice2020overfitting} and FGSM-PGI~\cite{jia2022prior}. The results are shown in Table~\ref{table:ImageNet}. When compared with PGD-AT, the  proposed FGSM-PGK achieves higher clean and robust accuracy. When compared with Free-AT, FGSM-RS and FGSM-PGI, the proposed FGSM-PGK achieves the best clean and robust accuracy. In terms of training efficiency, similar phenomenons are observed on other databases, our FGSM-PGK can be 3.3 times
% faster than the advanced PGD-AT.

% \section{Performance Analysis}
% \noindent \textbf{Comparisons under different $\ell_{\infty}$ distance metric.}  
\begin{table}[t]
\caption{  Ablation study of the proposed method. Robust accuracy (\%) and training time (hour)
using ResNet18 on the CIFAR-10 dataset on the best checkpoint. Number in bold indicates the best. 
}
\label{table:ablation}
  \scalebox{0.75}{
\begin{tabular}{@{}ccc|c|c|c|c|c@{}}
\toprule
\multicolumn{3}{c|}{Prior-Guided Knowledge}                                                   & \multirow{2}{*}{Clean} & \multirow{2}{*}{PGD-50} & \multirow{2}{*}{C\&W} & \multirow{2}{*}{AA} & \multirow{2}{*}{Time (h)} \\ \cmidrule(r){1-3}
\multicolumn{1}{c|}{Initialization} & \multicolumn{1}{c|}{Regularizer} & Weights &                        &                         &                       &                     &                           \\ \midrule
\multicolumn{1}{c|}{}              & \multicolumn{1}{c|}{}               &         & 83.82                  & 0.02                    & 0.00                 & 0.00               & 0.9                       \\ \midrule \midrule

\multicolumn{1}{c|}{\checkmark}              & \multicolumn{1}{c|}{}               &         & \textbf{84.92}                  & 47.5                    & 47.42                 & 44.92               & 0.9                       \\ \midrule
\multicolumn{1}{c|}{\checkmark}              & \multicolumn{1}{c|}{\checkmark}              &         & 81.77                  & 54.55                   & 51.08                 & 49.02               & 1.2                       \\ \midrule
\multicolumn{1}{c|}{\checkmark}              & \multicolumn{1}{c|}{}               & \checkmark       & 84.07                  & 50.48                   & 48.67                 & 46.08               & 0.9                       \\ \midrule
\multicolumn{1}{c|}{\checkmark}              & \multicolumn{1}{c|}{\checkmark}              & \checkmark       & 81.58                  & \textbf{55.31}          & \textbf{51.17}        & \textbf{49.51}      & 1.2                       \\ \bottomrule
\end{tabular}
}
\end{table}

\subsection{Ablation Study}
\label{Ablation_Study}
In our proposed fast adversarial training method, we propose three prior-guided knowledge, \emph{i.e.,} prior-guided initialization, regularizer, and weights. To evaluate the effectiveness of each prior-guided element in the proposed method, the ablation study with ResNet18 is conducted on CIFAR-10. We adopt a series of widely used robust evaluation methods,  which includes PGD-50, C\&W, and AA, to evaluate the adversarial robustness of all trained models. The results are shown in Table~\ref{table:ablation}. We also report the clean accuracy and training time of each model. 
Analyses are summarized as follows. First, when only incorporating the prior-guided initialization, it can prevent catastrophic overfitting and achieve limited robustness improvement. When incorporating the prior-guided initialization and regularizer, the robust performance significantly improves under all attack scenarios while it requires more training time. When incorporating the prior-guided initialization and weights, it requires the same training time as the model trained only with the prior-guided initialization and achieves better adversarial robustness. The results indicate that using the proposed prior-guided knowledge can contribute to preventing catastrophic overfitting and improving adversarial robustness with less training costs. Second, using all prior-guided knowledge could achieve the best adversarial robustness performance with little extra training time. It indicates that the prior-guided knowledge is compatible and combining them could achieve the best adversarial robustness performance.

\subsection{Performance Analysis}
\label{Performance_Analysis}
To investigate the effectiveness of our proposed method under different attack strengths used for training and evaluation, we follow the default settings of  \cite{wong2020fast,andriushchenko2020understanding} and adopt various fast adversarial training methods on CIFAR-10 with ResNet-18, trained and evaluated with different $\ell_{\infty}$ attack strengths. The results are shown in Fig.~\ref{fig:attack_strength}. It can be observed that previous fast adversarial training methods, such as GAT and NuAT, suffer from catastrophic overfitting when faced with larger $\ell_{\infty}$ attacks. Specifically, the previous state-of-the-art fast adversarial training method, NuAT, suffers from catastrophic overfitting when the $\ell_{\infty}$ attack strength is set to 14/255. In contrast, our proposed FGSM-PGK not only prevents catastrophic overfitting but also achieves significant improvements in adversarial robustness. It indicates
that using the prior-guided knowledge from the historical training process could improve the quality of adversarial examples and promote the model's adversarial robustness. 
\par  Previous works~\cite{zhao2020bridging,xu2019understanding,rebuffi2021fixing,li2023understanding} have demonstrated that a flatter adversarial loss profile indicates that the model is more robust to adversarial attacks. Hence, we analyze the adversarial loss landscapes of different fast adversarial training methods. To generate a loss landscape, we vary the network input along the linear space defined by a random Rademacher direction ($v$ direction) and an adversarial direction ($u$ direction) found by the PGD-50 attack. The results are shown in Fig.~\ref{fig:Fig_loss_landscape}.  Compared to previous fast adversarial training methods, the cross-entropy loss of the proposed FGSM-PGK method shows greater linearity in the adversarial direction. This suggests that incorporating the proposed prior-guided knowledge can better preserve the local linearity of the target model.

% To investigate the effectiveness of the proposed method on the different attack strength used for training and evaluation, following the default settings of \cite{wong2020fast,andriushchenko2020understanding}, we adopt different fast adversarial trainin methods on CIFAR-10 with ResNet-18 trained and evaluated with different attack strength $\ell_{\infty}$. The results is shown in Fig.~\ref{fig:attack_strength}. It can be observed that the previous fast adversarial training methods (such as GAT, NuAT, etc) also suffers from catastrophic overfitting under the larger attack strength $\ell_{\infty}$. However, the proposed FGSM-LAW not only prevents catastrophic overfitting but also achieve significant adversarial robustness improvement. 
\section{Conclusion}
\label{sec:conclusion}
In this paper, we investigate the differences in robust evolution between standard adversarial training and fast adversarial training processes across various networks and datasets, and we find that catastrophic overfitting could be related to the quality of adversarial examples used for training. Specifically, we observe that as the attack success rate of adversarial examples used for training deteriorates, the original fast adversarial training encounters catastrophic overfitting. To address this issue, we propose to adopt three prior-guided knowledge strategies: prior-guided initialization, regularizer, and model weights, to improve adversarial robustness with less extra training costs. Firstly, we propose to use the prior-guided initialization generated by using high-quality adversarial perturbations from the historical training
process. We also provide a theoretical analysis of the proposed initialization. Secondly, we propose a prior-guided regularization method using prior-guided and adversarial perturbations to improve the smoothness of the loss function around samples and ensure the model's robustness to both types of adversarial perturbations. Lastly, to further improve adversarial robustness, we propose a prior-guided weight averaging method for fast adversarial training, which dynamically averages the different model
weights of historical models using different decay rates. Extensive experimental evaluations conducted on four benchmark databases show that using prior-guided knowledge for fast adversarial training not only prevents catastrophic overfitting but also outperforms state-of-the-art fast adversarial training methods and several advanced standard adversarial training methods.


\bibliographystyle{plain}
\bibliography{refer}
\begin{IEEEbiography}[{\includegraphics[width=0.8in,height=1in,clip,keepaspectratio]{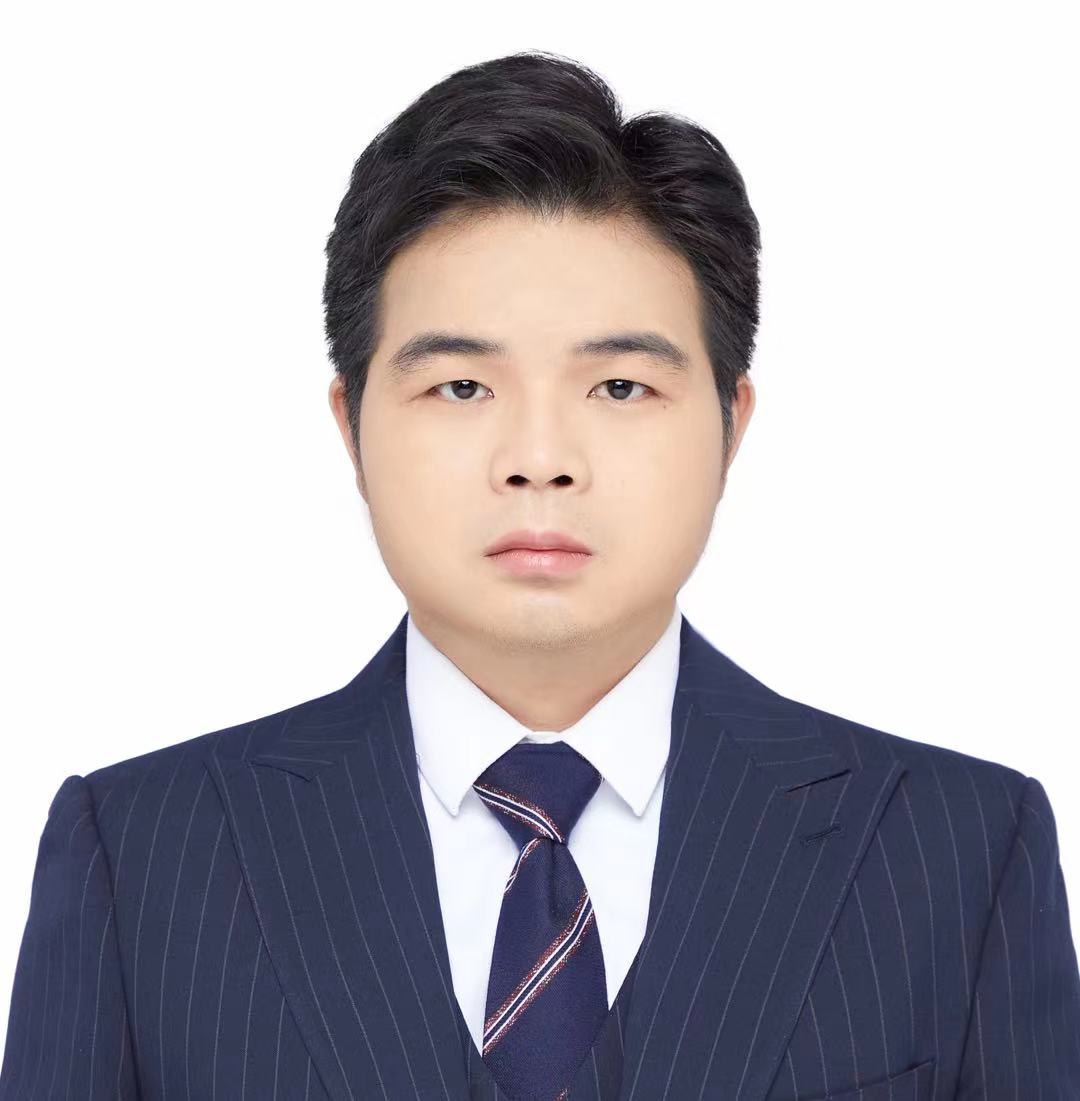}}]
{Xiaojun Jia} 
received his B.S. degree in Software Engineering from China University of Geosciences, China. He is now a Ph.D student in State Key Laboratory of Information Security, Institute
of Information Engineering, Chinese Academy of Sciences and School of
Cyber Security, University of Chinese Academy of Sciences, Beijing. His research interests include computer vision, deep learning and adversarial machine learning. He is the author of referred journals and conferences in IEEE CVPR, AAAI, ACM Multimedia etc.
\end{IEEEbiography}
\vspace{-6mm}
\begin{IEEEbiography}
[{\includegraphics[width=0.8in,height=1in,clip,keepaspectratio]{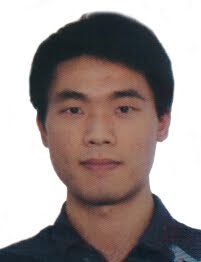}}]
{Yong Zhang} 
received the Ph.D. degree  in pattern recognition and intelligent systems from the Institute of Automation, Chinese Academy of Sciences in 2018. From 2015 to 2017, he was a Visiting Scholar with the Rensselaer Polytechnic Institute. He is currently with the Tencent AI Lab. His research interests include computer vision and machine learning.
\end{IEEEbiography}

% \vspace{-6mm}
% \begin{IEEEbiography}
% [{\includegraphics[width=0.8in,height=1in,clip,keepaspectratio]{photo/Yang_Bai}}]
% {Yang Bai} 
% received the B.S. degree from Department of Electronic Engineering, Tsinghua University in 2017. She is currently a Ph.D student in Tsinghua University. Her research interests include computer vision and machine learning, especially adversarial or backdoor robustness.
% \end{IEEEbiography}

\begin{IEEEbiography}[{\includegraphics[width=1in,height=1.25in,clip,keepaspectratio]{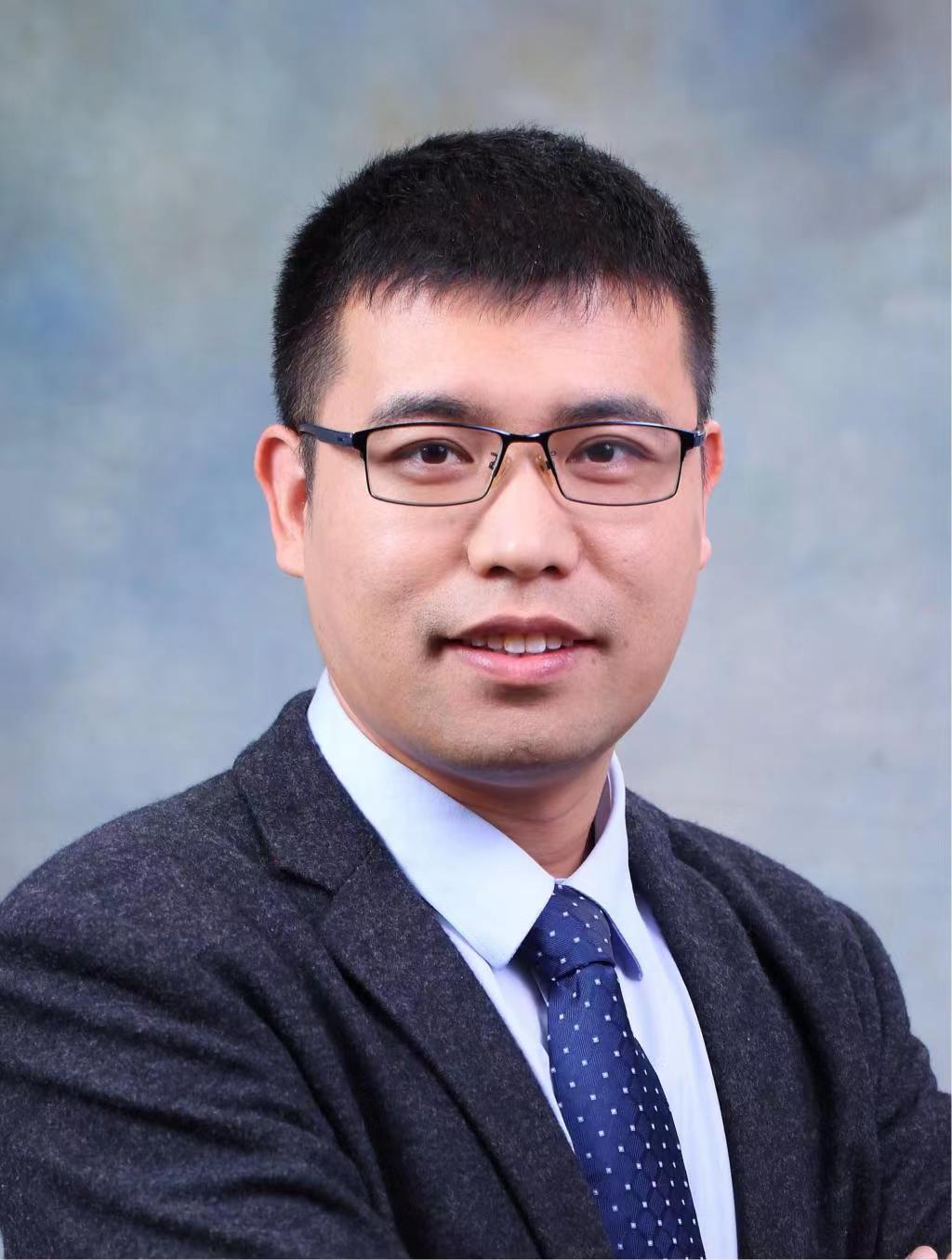}}]{Xingxing Wei}
received his Ph.D degree in computer science from Tianjin University, and B.S. degree in Automation from Beihang University, China. He is now an Associate Professor in Beihang University (BUAA). His research interests include computer vision, adversarial machine learning and its applications to multimedia content analysis. He is the author of referred journals and conferences in IEEE TPAMI, TMM, TCYB, TGRS, IJCV, PR, CVIU,  CVPR, ICCV, ECCV, ACMMM, AAAI, IJCAI etc.
\end{IEEEbiography}

\begin{IEEEbiography}
[{\includegraphics[width=0.8in,height=1in,clip,keepaspectratio]{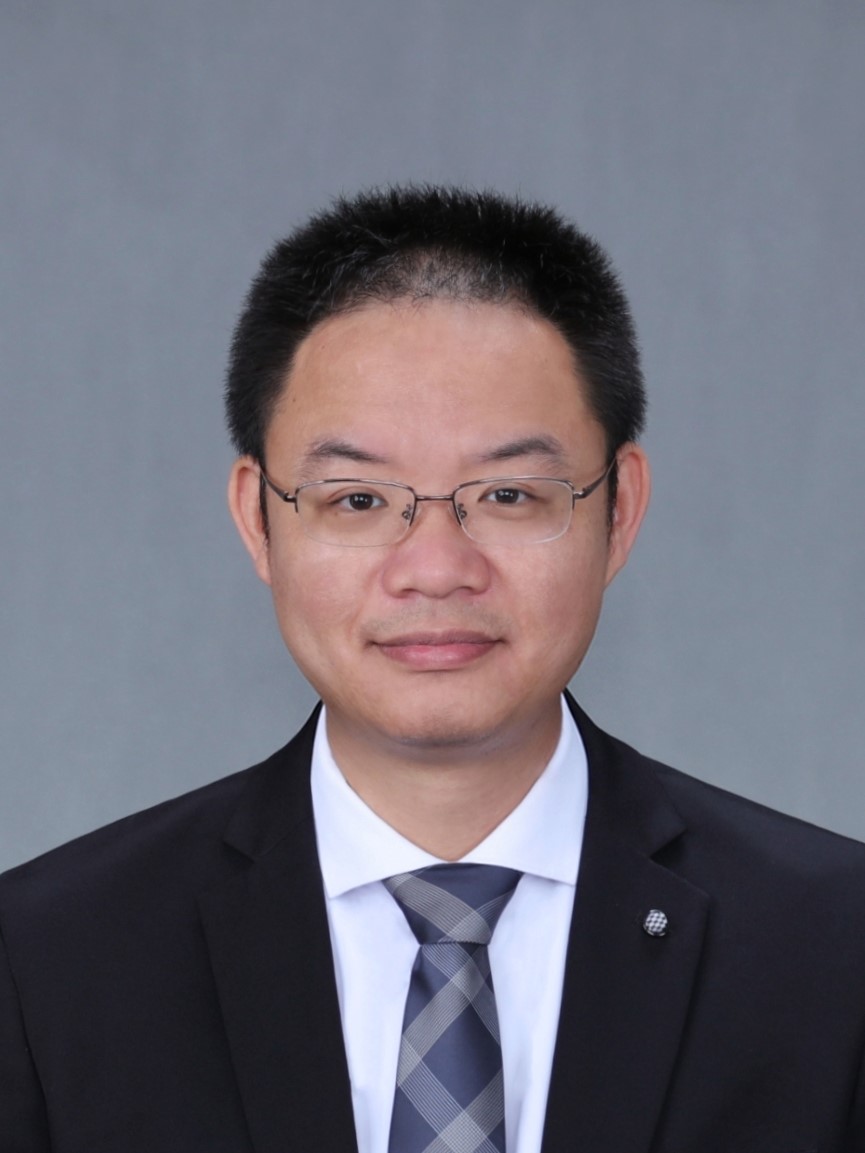}}]
{Baoyuan Wu} is an Associate Professor of School of Data Science, the Chinese University of Hong Kong, Shenzhen (CUHK-Shenzhen). He is also the director of the Secure Computing Lab of Big Data, Shenzhen Research Institute of Big Data (SBRID). On June 2014, he received the PhD degree from the National Laboratory of Pattern Recognition, Institute of Automation, Chinese Academy of Sciences. From November 2016 to August 2020, he was a Senior and Principal Researcher at Tencent AI lab. His research interests are AI security and privacy, machine learning, computer vision and optimization. He has published 40+ top-tier conference and journal papers, including TPAMI, IJCV, NeurIPS, CVPR, ICCV, ECCV, ICLR, AAAI, and one paper was selected as the Best Paper Finalist of CVPR 2019. He serves as an Associate Editor of Neurocomputing, Area Chair of ICLR 2022, AAAI 2022 and ICIG 2021, Senior Program Committee Member of AAAI 2021 and IJCAI 2020/2021, Task Force Member of CCF and CAA. He is the principal investigator of General Program of National Natural Science Foundation of China, 2021 CCF-Tencent Rhino-Bird Young Faculty Open Research Fund, and 2021 Tencent Rhino-Bird Special Research Fund.
\end{IEEEbiography}

\begin{IEEEbiography}[{\includegraphics[width=1in,height=1.25in,clip,keepaspectratio]{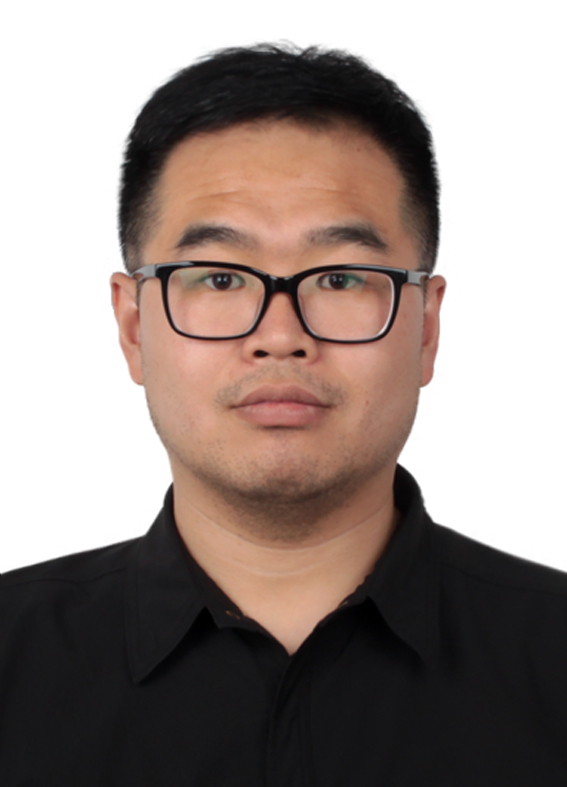}}]{Ke Ma} is an associate professor with the School of Electronic, Electrical and Communication Engineering, University of Chinese Academy of Sciences (UCAS), Beijing, China. He received the B.S. degree in mathematics from Tianjin University in 2009, M.E. degree in software engineering from Beihang University (BUAA) in 2013, and the Ph.D. degree in computer science from the Key Laboratory of Information Security (SKLOIS), Institute of Information Engineering (IIE), Chinese Academy of Sciences (CAS), in 2019. His research interests include rank aggregation and algorithmic game theory.
\end{IEEEbiography}

\begin{IEEEbiography}
[{\includegraphics[width=0.8in,height=1in,clip,keepaspectratio]{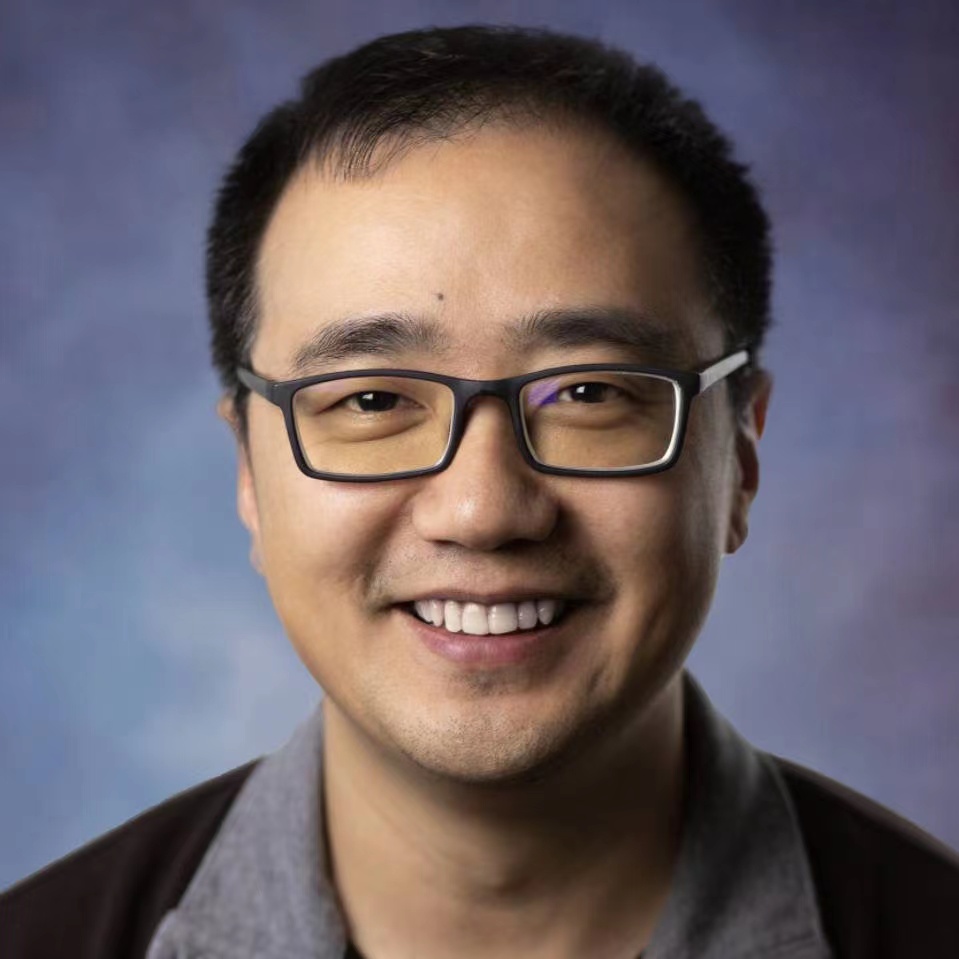}}]
{Jue Wang} received the B.E. and M.Sc. degrees from the Department of Automation, Tsinghua University, Beijing, China, and the Ph.D. degree in Electrical Engineering from the University of Washington in Seattle WA, USA. He is currently the Director of the Visual Computing Center at Tecent AI Lab. He was Senior Director at Megvii Research from 2017 to 2020. He was Principle Research Scientist at Adobe Research from 2007 to 2017. He has published more than 140 peer-reviewed research articles in the areas of Computer Vision, Computer Graphics and HCI,  and holds more than 60 international patents. He was a co-organizer for IEEE International Conference on Multimedia and Expo 2016 and IEEE International Conference on Computational Photography 2012.  He is Associated Editor for IEEE Transactions on Pattern Analysis and Machine Intelligence (T-PAMI), The Visual Computer and International Journal of Computer Games Technology. He was the recipient of the Microsoft Research Fellowship and the Yang Research Award of the University of Washington in 2006. He is a senior member of IEEE and ACM.
\end{IEEEbiography}

\begin{IEEEbiography}
[{\includegraphics[width=0.8in,height=1in,clip,keepaspectratio]{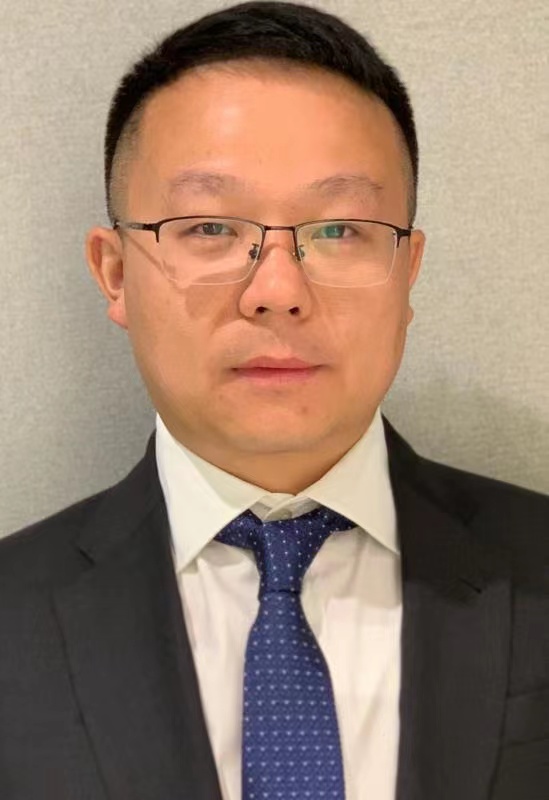}}]
{Xiaochun Cao}(SM'14)
received the B.S. and M.S. degrees in computer science from Beihang University, Beijing, China, and the Ph.D. degree in computer science from the University of Central Florida, Orlando, FL, USA. After graduation, he spent about three years at ObjectVideo Inc. as a Research Scientist. He is with School of Cyber Science and Technology, Shenzhen Campus, Sun Yat-sen University, Shenzhen 518107, P.R. China. He has authored and coauthored more than 100 journal and conference papers.
Prof. Cao is a Fellow of the IET. He is on the Editorial Boards of the IEEE Transactions on Image Processing, IEEE Transactions on Multimedia, IEEE Transactions on Circuits and Systems for Video Technology. His dissertation was nominated for the University of Central Florida's university-level Outstanding Dissertation Award. In 2004 and 2010, he was the recipient of the Piero Zamperoni Best Student Paper Award at the International Conference on Pattern Recognition.
\end{IEEEbiography}

\end{document}